\pdfoutput=1
\documentclass[11pt]{article}

\usepackage{ACL2023}

\usepackage{times}
\usepackage{latexsym}
\usepackage{microtype}
\usepackage{stmaryrd}
\usepackage{booktabs}
\usepackage{multirow}
\usepackage{graphicx}
\usepackage{caption}
\usepackage{subcaption}
\usepackage{bm}
\usepackage{algorithmicx,algorithm}
\usepackage[noend]{algpseudocode}
\usepackage{amsthm}
\usepackage{enumitem}
\usepackage{amssymb}
\usepackage{xspace}
\usepackage{mathtools}
\usepackage{comment}
\usepackage{dsfont}

\usepackage[T1]{fontenc}
\usepackage[utf8]{inputenc}

\usepackage{inconsolata}

\usepackage{amsmath,amsfonts,bm}

\def\eqref#1{equation~\ref{#1}}
\def\1{\bm{1}}

\DeclareMathAlphabet{\mathsfit}{\encodingdefault}{\sfdefault}{m}{sl}
\SetMathAlphabet{\mathsfit}{bold}{\encodingdefault}{\sfdefault}{bx}{n}

\def\gU{{\mathcal{U}}}

\DeclareMathOperator*{\argmax}{arg\,max}

\definecolor{darkred}{HTML}{b30000}
\definecolor{darkgreen}{HTML}{006400}
\definecolor{lightgreen}{HTML}{1d7b21}
\definecolor{lightred}{HTML}{9d0000}
\definecolor{verylightgreen}{HTML}{00bb00}

\newtheorem*{definition*}{Definition}

\newtheorem*{assumption*}{Assumption}

\renewcommand{\sectionautorefname}{\S\kern-0.2em}
\renewcommand{\subsectionautorefname}{\S\kern-0.2em}

\renewcommand{\vec}[1]{\boldsymbol{#1}}

\newcommand{\instr}{\vec u}
\newcommand{\traj}{\vec e}

\newcommand{\humantraj}{\traj^{\textrm{h}}}

\newcommand{\agenttraj}{\traj^{\star}}

\newcommand{\pragspeaker}{S_{\textrm{prag}}}
\newcommand{\basespeaker}{S_{\textrm{base}}}
\newcommand{\tomlistener}{L_{\textrm{ToM}}}
\newcommand{\humanlistener}{L_{h}}
\newcommand{\agentspeaker}{S_{r}}

\newcommand{\agentinstr}{\hat{\vec \instr}}

\newcommand{\evalmetric}{\Psi}
\newcommand{\perf}{\rho}

\newcommand{\ppgsearch}{\textrm{PPG}_{\textrm{search}}}
\newcommand{\ppgprag}{\textrm{PPG}_{\textrm{pragmatic}}}

\newcommand{\sr}{\textsc{SR}\xspace}
\newcommand{\spl}{\textsc{SPL}\xspace}
\newcommand{\ndtw}{\textsc{NDTW}\xspace}
\newcommand{\sdtw}{\textsc{SDTW}\xspace}

\setlist{nosep}

\newcommand{\gpt}{GPT-2\xspace}
\newcommand{\encdeclstm}{EncDec-LSTM\xspace}
\newcommand{\encdectrans}{EncDec-Transformer\xspace}

\def\CircleArrowright{\ensuremath{%
  \rotatebox[origin=c]{310}{$\circlearrowright$}}}
\newcommand{\recurrentbert}{VLN$\protect\CircleArrowright$BERT\xspace}
\newcommand{\vlnbert}{VLN-BERT\xspace}
\newcommand{\envdrop}{EnvDrop-CLIP\xspace}

\usepackage{tikz}

\newcommand{\noneperf}[1]{\scriptsize\textcolor{gray}{($\blacktriangle$ #1)}}
\newcommand{\noneperfdown}[1]{\scriptsize\textcolor{gray}{($\blacktriangledown$ #1)}}
\newcommand{\upperf}[1]{\scriptsize\textcolor{lightgreen}{($\blacktriangle$ #1)}}
\newcommand{\downperf}[1]{\scriptsize\textcolor{lightred}{($\blacktriangledown$ #1)}}

\usepackage{etoolbox}
\newcommand{\zerodisplayskips}{%
  \setlength{\abovedisplayskip}{5pt}%
  \setlength{\belowdisplayskip}{5pt}%
  \setlength{\abovedisplayshortskip}{0pt}%
  \setlength{\belowdisplayshortskip}{0pt}}
\appto{\normalsize}{\zerodisplayskips}
\appto{\small}{\zerodisplayskips}
\appto{\footnotesize}{\zerodisplayskips}

\title{Define, Evaluate, and Improve Task-Oriented \\ Cognitive Capabilities for Instruction Generation Models}

\author{$^{\spadesuit}$Lingjun Zhao\Thanks{ The first two authors contribute equally. } \ \and
  $^{\clubsuit}$Khanh Nguyen\footnotemark[1]  \ \and $^{\spadesuit\diamondsuit}$Hal Daum\'e III \\
  $^{\spadesuit}$University of Maryland--College Park \ \ $^{\clubsuit}$Princeton University \ \
  $^{\diamondsuit}$Microsoft Research \\
  \texttt{lzhao123@umd.edu } \\}

\begin{document}
\maketitle

\begin{abstract}
Recent work studies the cognitive capabilities of language models through psychological tests designed for humans. 
While these studies are helpful for understanding the general capabilities of these models, there is no guarantee that a model possessing sufficient capabilities to pass those tests would actually \textit{use} those capabilities in performing real-life tasks.
In this work, we formulate \textit{task-oriented} cognitive capabilities, which are human-like cognitive capabilities that language models leverage to perform tasks. 
These capabilities are (i) the ability to quickly generate good candidate utterances (the search capability) (ii) the ability to predict how a listener interprets those utterances and choose the most appropriate one (the pragmatic capability).
We design an evaluation scheme for comparing these capabilities of a language model with those of a human.
Applying this scheme to examine various models in a navigation instruction generation problem, we find that their pragmatic capability is severely lacking.
This insight leads us to augment them with better models of the listener and obtain a significant boost of 11\% in success rate in guiding real humans.
Our work advocates for having a principled procedure for aligning language models with humans that involves (i) formulating task-oriented capabilities, (ii) devising a method to quantify their deficiency, and (iii) iteratively improving them.
\looseness=-1
\end{abstract}

\section{Introduction}

To communicate successfully with humans, language models must possess cognitive capabilities similar to those that facilitate human communication.
Examining the cognitive capabilities of language models is notoriously challenging because the operations of these models are largely unintelligible to humans. 
Psychologists faced similar challenges when investigating human cognition, and have devised various behavioral tests to diagnose human cognitive capabilities \citep{premack1978does,wimmer1983beliefs,baron1985does,gopnik1988children}. 
Recent work \citep{sap2022neural,kosinski2023theory,ullman2023large} applies these tests to evaluate large language models by inputting the tests to these models as prompts and verifying whether they behave like a normal human would. 
\looseness=-1

While this approach is helpful for understanding the general limitations of language models, it has two potential drawbacks.
First, it is applicable to only large language models that can comprehend human-written prompts, entangling linguistic capability with reasoning capability. 
Second, it shows that a language model can or cannot demonstrate certain mental skills, but does not imply that the model would employ those skills to perform a downstream task.
For example, passing false-belief tests does not guarantee that a model will reason about the interpretation of the readers when generating summaries.
In general, scoring high on psychological tests may not be sufficient to ensure language models would behave like humans in real-life scenarios. 
\looseness=-1

\begin{figure}[t!]
    \centering
    \includegraphics[width=0.95\linewidth]{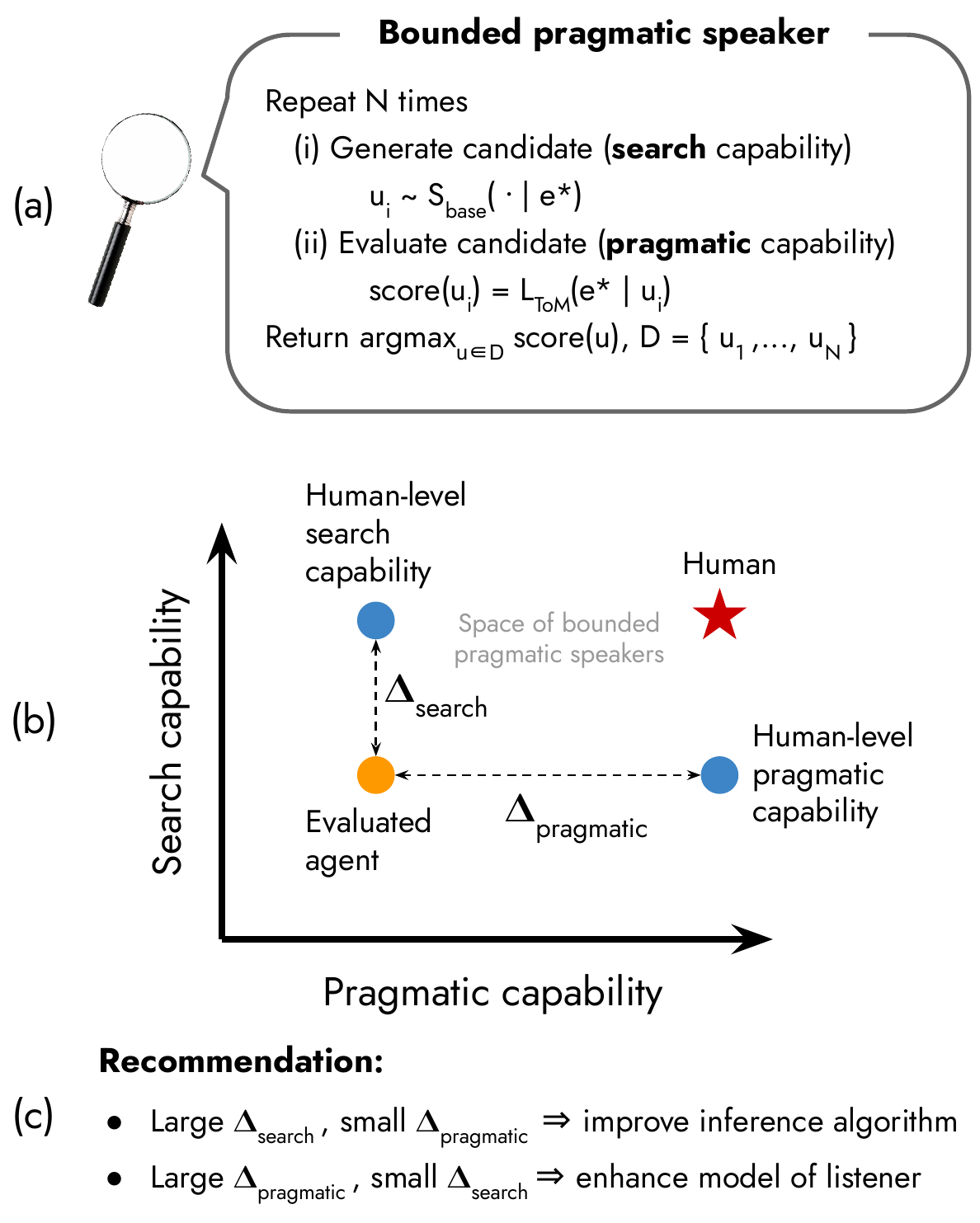}
    \caption{We propose a framework called \textit{bounded pragmatic speaker} which can characterize pragmatic reasoning in both humans and language models (a). A bounded pragmatic speaker is composed of a base speaker $\basespeaker$, representing prior knowledge that helps generate instructions efficiently, and a theory-of-mind (ToM) listener $\tomlistener$, a hypothetical model of how the real listener interprets instructions. Viewing language models and humans through this unifying lens enables comparing their cognitive capabilities (b). To evaluate a capability of a model, we compare it with an oracle model which is identical except that the evaluated capability is at human level. The outcome of our evaluation can better inform the future direction for improving the model (c).  \looseness=-1}
    \label{fig:overview}
\end{figure}

In this work, we take a different approach to evaluating the cognitive capabilities of language models. 
We define and evaluate \textit{task-oriented} cognitive capabilities, which are human-like capabilities that a model actually employs to perform the task it is designed for. 
Enhancing these capabilities thus warrants improved performance on the task.
To identify these capabilities, we build on two lines of work from socio-cognitive science: Bayesian models of cooperative communication \cite{wang2020mathematical,goodman2016pragmatic,shafto2014rational} and studies on how humans implement Bayesian reasoning \cite{sanborn2016bayesian,sanborn2010rational,vul2014one,mamassian2002bayesian}. 
We propose a mathematical cognitive model called \textit{bounded pragmatic speaker}, which can reasonably characterize the reasoning processes of both humans and language models. 
Casting humans and language models in the same way enables us to juxtapose their cognitive capabilities. 
We mathematically formulate two capabilities that a bounded pragmatic agent must possess in order to generate optimally pragmatic utterances.  
These conditions correspond to well-known cognitive capabilities of humans: (i) the ability to efficiently generate relevant utterances (the \textit{search} capability) \cite{bloom1980completion,gold2000deriving,trosborg2010pragmatics} and (ii) the ability to accurately simulate the listener's interpretations of their utterances (the \textit{pragmatic} capability) \citep{premack1978does,gopnik1988children,tomasello2019becoming,call2011does, frank2012predicting}.
We design a simple procedure to quantitatively evaluate these capabilities of a language model.
To evaluate each capability, we compute the task performance gap between the model and an \textit{oracle} model, which is identical except that the evaluated capability of this model is at human level. 
\autoref{fig:overview} illustrates our procedure, which theoretically can be applied to any language model.

We evaluate various language models on a navigation instruction generation problem \cite{anderson2018vision}, where a model generates English instructions to guide real humans in photo-realistic 3D environments.\footnote{Our human-evaluation dataset and interface are publicly released at \url{https://lingjunzhao.github.io/coop_instruction.html}.}
Our evaluation reveals an interesting finding: all evaluated agents possess relatively efficient search capability but inadequate pragmatic capability. 
We improve the pragmatic capability of the evaluated models by enabling them to reason probabilistically about human listeners \cite{andreas2016reasoning,fried2017unified}, employing state-of-the-art instruction-following agents \cite{magalhaes2019effective,shen2021much,hong2021vln} as models of human listeners. 
We obtain significant improvement in success rate over the original agents, shrinking the gap with human performance on held-out data by 36\%. 
Towards eliminating the remaining gap, we illustrate with empirical evidence a major challenge in developing better listener models.
Specifically, when the instruction-following agents are employed as listener models for the instruction-generating agent, they are required to evaluate \textit{AI-generated} instructions, which may be significantly different from human-generated instructions. 
Hence, a standard supervised-learning training scheme that only exposes these models to human-generated instructions would be inadequate for learning reliable models. 
We thus call for construction of novel datasets, algorithms, and evaluation methods for developing the pragmatic capability of language models.
\looseness=-1

\section{Related Work}

\paragraph{Navigation Instruction Generation.} 

Instruction generation has been commonly studied in navigation settings \cite{anderson1991hcrc,byron2010report,koller2010report,striegnitz2011report,goeddel2012dart,fried2017unified,fried2018speaker}.
The Matterport3D simulator and the accompanying datasets (R2R \cite{anderson2018vision}, R4R \cite{jain2019stay}, and RxR \cite{ku2020room}) offer more challenging settings by combining photo-realistic scenes with long, verbally rich instructions. 
Recent work on evaluating instruction generation agents \cite{zhao2021evaluation} reveals the ineffectiveness of standard learning and modeling approaches to this problem. 
\citet{wang2021less} improve the accuracy and interpretability of instructions in the RxR setting. 
\citet{kamath2022new} leverage this model to synthesize additional data for training instruction-following agents.
Our work aim to offer useful principles to further improve these models. 

\paragraph{Mathematical Models of Human Communication.}
Different from communication within agents \cite{lazaridou2020multi, roman2020rmm}, human communication is a cooperative act \cite{grice1975logic,scott2014speaking,tomasello2019becoming}. 
Pragmatic communication in humans may involve different cognitive capabilities like basic understanding of language and social rules \cite{trosborg2010pragmatics} and reasoning about the physical world \citep{bender2020climbing} and human behavior \cite{enrici2019theory,rubio2021pragmatic}.
Our work describes similar capabilities but provides a formal mathematical description.  
Development of mathematical models of human communication have been greatly useful for understanding human behaviors \cite{ho2016showing,sumers2022talk} and building communication agents \cite{andreas2016reasoning,fried2017unified,fried2018speaker,meta2022human,lin2022inferring, zhu2021few, bao2022learning}. 
\citet{wang2020mathematical} unify these models under an optimal-transport framework.
The model we propose in this work is a generalized version capturing the essence of these models. 

\paragraph{Evaluating Cognitive Capabilities of Neural Networks.} 
 Many benchmarks for evaluating the cognitive capabilities of AI-based agents have been created, focusing on theory-of-mind capabilities \cite{le2019revisiting, nematzadeh-etal-2018-evaluating}, grounding \cite{lachmy2021draw,udagawa2019natural, haber2019photobook}, or commonsense reasoning \cite{talmor2018commonsenseqa,levesque2012winograd,zellers2019recognition,sap-etal-2019-social}. 
Large language models have demonstrated exceptional performance on following human instructions and solving complex reasoning tasks \citep{bubeck2023sparks,anil2023palm}, raising the question of whether their cognitive capabilities are similar or as advanced as those of humans.
\citet{mahowald2023dissociating} advocate for separating formal competence (knowledge about linguistic rules and patterns) from their functional competence (knowledge about the world usage in the world) when assessing these models. 
Our bounded pragmatic speaker framework mathematically formalizes this description, allowing for quantitative evaluation of these competencies. 
Recent work \cite{sap2022neural,kosinski2023theory,ullman2023large,hu2022fine} examines cognitive capabilities of large language models through tests inspired by human psychological tests.
The goal of these studies is to determine the limits of large language models, potentially calibrating the expectation on them. 
On the other hand, our focus is to devise a method that can be applied to language models of any size and benchmark cognitive capabilities that are relevant for accomplishing a specific task. 

\looseness=-1

\section{Problem Setting}
\label{sec:problem}

We are concerned with instruction generation: learning a speaker agent $r$ that generates language instructions to guide a human listener $h$ to reach states in an environment. 

\paragraph{Human Listener.}
We imagine a human listener $h$ acting in a partially observed environment with states $s$.
The human does not have access to states but only observations $o^h$ and takes actions $a^h$.
An \textit{instruction} $\instr \in \mathcal{U}$ is a language utterance consisting of words.
A \textit{trajectory} $\traj = (s_1, o_1, a_1, \cdots, s_T, o_T, a_T)$ is an execution of an instruction.
The human can follow instructions to generate trajectories in the environment. 
For example, in an indoor navigation setting, upon hearing ``\textit{go the kitchen and stop next to the oven}'', a human walks to the specified location. 
We define $\humanlistener(\traj \mid \instr)$ as the probability that the human generates $\traj$ upon hearing $\instr$. \looseness=-1

\paragraph{Speaker Agent.}

In each task, the speaker agent first imagines an \textit{intended trajectory} $\agenttraj = (s_1, o_1^r, a_1^r, \cdots, s_T, o_T^r, a_T^r)$, which specifies a path to get to an intended goal state $s_T$ from the human's current state $s_1$.
Because the human's actions and perception may differ from those of the speaker, they may not be able to comprehend $\agenttraj$ even if it is presented to them.  
Thus, the speaker needs to translate the trajectory into an instruction $\agentinstr$ that the human can understand and follow.
To do so, it implements a \textit{language model} $\agentspeaker(\instr \mid \traj)$, and an \textit{inference algorithm} $\textrm{Gen}(\agentspeaker, \traj)$ to craft instructions based on the model (e.g., greedy or beam-search decoding).
The speaker's objective is to generate instructions that maximize the expected chance of the listener reconstructing the intended trajectories \looseness=-1
\begin{align}
    &\argmax_{\agentspeaker} \mathbb{E}_{\agenttraj} \left[ \humanlistener(\agenttraj \mid \textrm{Gen}(\agentspeaker, \agenttraj)) \right]  
\label{eqn:objective}
\end{align} 

\paragraph{Evaluation.}
The speaker is evaluated using a dataset $\mathcal{D}_{\textrm{eval}}$ of held-out trajectories.
For each trajectory $\agenttraj_k \in \mathcal{D}_{\textrm{eval}}$, we generate an instruction $\agentinstr_{k} = \textsc{Gen}(\agentspeaker, \agenttraj_k)$.
The instruction is then presented to a (real) human listener to follow, producing a trajectory $\humantraj_k \sim \humanlistener(\cdot \mid \agentinstr_k)$.
The performance of the speaker, denoted by $\rho(r)$, is the average similarity, $\Psi$, between the human-generated and the intended trajectories:
\begin{align}
    \perf(r) \triangleq \frac{1}{|\mathcal{D}_{\textrm{eval}}|}\sum_{\agenttraj_k  \in \mathcal{D}_{\textrm{eval}}} \evalmetric(\humantraj_k, \agenttraj_k)  
\label{eqn:perf_metric}
\end{align} We will specify the choices for the metric $\Psi$ in the experimental setup section (\autoref{sec:exp_setup}).
\section{Task-Oriented Cognitive Capabilities}
\label{section_framework}

Human have evolved highly effective cognitive capabilities for communication.
How can we endow a speaker agent with similar capabilities and quantify the degree of resemblance between its capabilities and those of a human?

We propose a mathematical framework that reasonably characterizes the human cognitive process for instruction generation (\autoref{sec:math_model}).
We show that this model can also describe the operation of language models, which allows us to compare them with humans on specific cognitive capabilities.
We identify two capabilities that are requisite for any agent implementing our framework to generate optimal instructions (\autoref{sec:essential_capabilities}), and introduce an evaluation scheme for collating these capabilities (\autoref{sec:evaluation_scheme}). \looseness=-1  

\subsection{A Mathematical Cognitive Model of Instruction Generation}
\label{sec:math_model}

\begin{figure*}[t!]
\centering
\includegraphics[width=.9\linewidth]{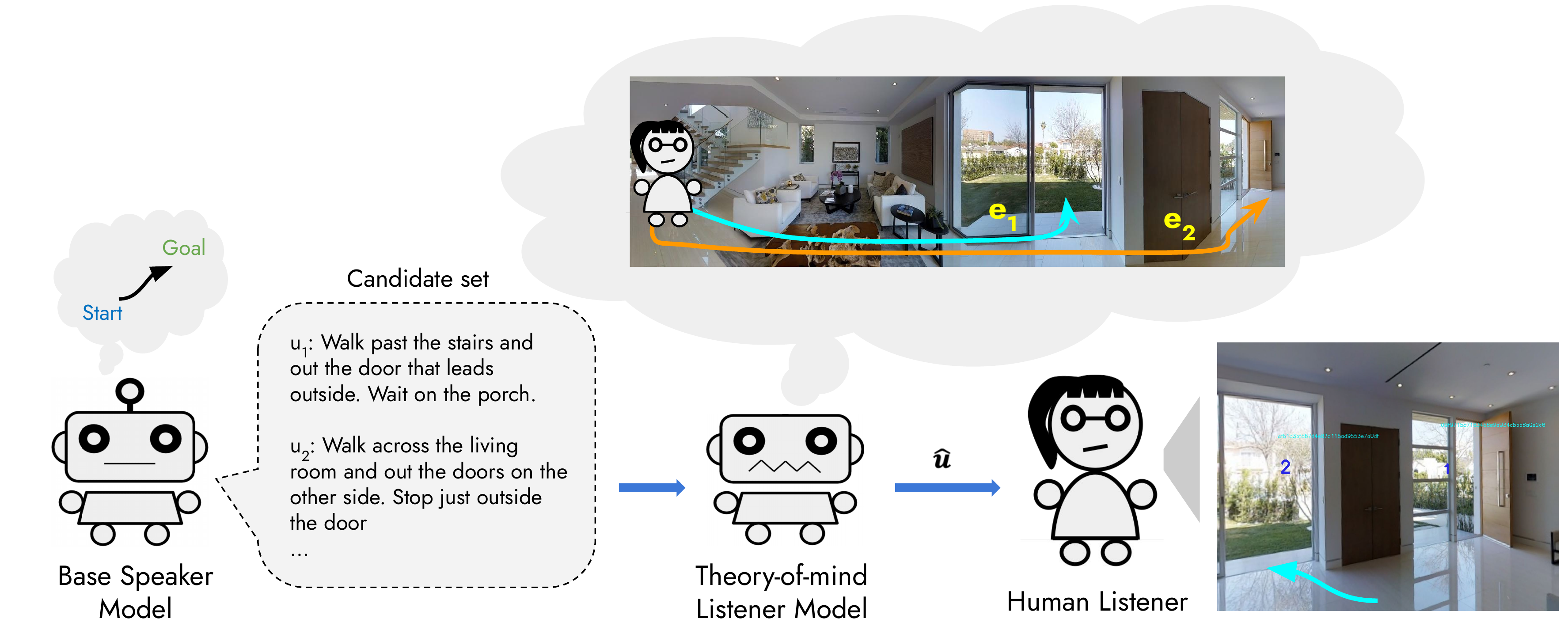}
\caption{The cognitive process of a bounded pragmatic speaker. In every task, the speaker first imagines a trajectory it wants to convey to the human listener. To reduce the search space, it then uses the \textit{base speaker} to generate a small set of relevant candidate instructions. After that, it employs the \textit{theory-of-mind listener} to simulate how the human listener would follow each instruction in the candidate set. The speaker finally elects the candidate instruction that causes the theory-of-mind listener to generate the trajectory most similar to the intended trajectory. The output instruction is finally sent to the human listener for a real execution in the environment.}
\label{fig:bounded_pragmatic}
\end{figure*}

To formulate how humans generate instructions, we build on mathematical models of cooperative communication \cite{wang2020mathematical,goodman2016pragmatic,shafto2014rational}.
We consider a general version where a speaker agent constructs a \textit{pragmatic speaker} model $\pragspeaker(\instr \mid \traj)$ based on two constituents: a \textit{base speaker} model $\basespeaker(\instr \mid \traj)$ and a \textit{theory-of-mind (ToM) listener} model $\tomlistener(\traj \mid \instr)$. 
The base speaker represents general knowledge of the agent about the world and the language it speaks. 
The ToM listener reflects situated knowledge about the listener, simulating how they would behave in the environment given an instruction. 
A pragmatic speaker aims to maximize the chance of the listener interpreting its instruction correctly, but it is still influenced by its general knowledge (e.g., social biases, language style).
Formally, it is defined as:
\begin{align}
    \pragspeaker(\instr \mid \traj) \propto  \tomlistener(\traj \mid \instr) \basespeaker(\instr \mid \traj)
\label{eqn:cooperative_speaker}
\end{align} 
To convey an intended trajectory $\agenttraj$, this speaker utters an instruction of maximum probability under its model:
\begin{align}
    \agentinstr_{\textrm{prag}} 
    &\triangleq \argmax_{\instr \in \mathcal{U}} \pragspeaker(\instr \mid \agenttraj) \nonumber \\
    &= \argmax_{\instr \in \mathcal{U}} \tomlistener(\agenttraj \mid \instr) \basespeaker(\instr \mid \agenttraj)
\label{eqn:pragmatic_selection}
\end{align} \looseness=-1

\paragraph{Humans as bounded pragmatic speakers.} The pragmatic speaker model accounts for human behaviors highly accurately on problems where $\mathcal{U}$ is a small discrete space \cite{frank2012predicting}.
However, in problems like instruction generation where $\mathcal{U}$ is an unbounded set of linguistic expressions, it is unlikely that humans, which are known to be agents with bounded rationality \cite{simon1957models}, are able to compute the optimal utterance in \autoref{eqn:pragmatic_selection} exactly.
A hypothesis, supported by empirical evidence, is that humans perform approximate inference via Monte-Carlo sampling \cite{sanborn2016bayesian,sanborn2010rational,vul2014one,mamassian2002bayesian}.
Applying this hypothesis to our setting, we derive a more practical model of how human generate instructions, in which they perform the search for the best utterance on only a subspace $\mathcal{U}_{\textrm{sub}}$ of $\mathcal{U}$ defined by a set of candidates sampled from $\basespeaker$
\begin{align}
    &\agentinstr_{\textrm{bounded-prag}} 
    \triangleq \argmax_{\instr \ \in \ \mathcal{U}_{\textrm{sub}} \ \subset \ \mathcal{U} } \tomlistener(\agenttraj \mid \instr) \label{eqn:approx_pragmatic} 
\end{align} where $\mathcal{U}_{\textrm{sub}} = \{ \instr_i \sim \basespeaker(\cdot \mid \agenttraj) \mid  1 \leq i \leq N \}$.
We call an agent that generates instructions according to \autoref{eqn:approx_pragmatic} a \textit{bounded pragmatic speaker} (\autoref{fig:bounded_pragmatic}). 
For such a speaker, instruction generation involves two tasks: candidate generation  (performed by $\basespeaker$) and candidate evaluation (performed by $\tomlistener$).   
The former task ensures that the generation of an instruction is efficient, while the latter guarantees the generated instruction conveys the intended meaning to the human listener.\looseness=-1

\subsection{Formulating Task-Oriented Cognitive Capabilities}
\label{sec:essential_capabilities}

What cognitive capabilities enable humans to generate effective instructions? Viewing humans as bounded pragmatic speakers allows us to mathematically characterize those capabilities.
Specifically, we require a bounded pragmatic speaker to be able to output the optimal utterance, i.e. satisfying
\begin{align}
    \agentinstr_{\textrm{bounded-prag}} = \instr^{\star} \triangleq \argmax_{\instr} \humanlistener(\agenttraj \mid \instr)
\label{eqn:pragmatic_condition}
\end{align} where $\humanlistener$ is the human listener. 

For this equation to hold, the constituent models $\basespeaker$ and $\tomlistener$ of the bounded pragmatic speaker must meet certain conditions.
The condition for $\basespeaker$ is that the candidate set it generates must contain the optimal instruction, i.e. $\instr^{\star} \in \mathcal{U}_{\textrm{sub}}$.
This condition requires $\basespeaker$ to be capable of quickly generating candidates and placing high probability on $\instr^{\star}$ so the instruction can be found by sampling a few candidates.  
We refer to this capability as the \textit{search capability}.

Meanwhile, the condition for $\tomlistener$ is that it has to rank $\instr^{\star}$ first among the candidates in $\mathcal{U}_{\textrm{sub}}$. 
Meeting this condition demands having the capability of constructing a mental emulation of the human listener and simulating the actions of the listener after receiving an instruction. 
We refer to this capability as the \textit{pragmatic capability}.

The search and pragmatic capabilities are orthogonal and complementary. 
An agent with flawless pragmatic capability can evaluate the goodness of instructions given to it, but may not be able to efficiently generate good instructions by itself. 
In contrast, an agent with effective search capability can quickly bring to attention highly relevant utterances but cannot select the best one to output if its ToM model is erroneous. 

\subsection{Evaluating Task-Oriented Cognitive Capabilities}
\label{sec:evaluation_scheme}

We have defined two cognitive capabilities that are requisite for humans in instruction generation.
In this section, we will prove that a language model can also be cast as a bounded pragmatic speaker.
Hence, we can compare it with a human on the two cognitive capabilities. 

\paragraph{Language models as bounded pragmatic speakers.}
We consider a speaker agent $r$ that learns a language model $\agentspeaker(\instr \mid \traj)$ and  runs an inference algorithm to compute an output $\agentinstr_{\textrm{infer}} = \text{GEN}(\agentspeaker, \agenttraj) \approx \argmax_{\instr \in \mathcal{U}} \agentspeaker(\instr \mid \agenttraj)$.
Generative LSTM- or Transformer-based models that implement greedy or beam-search decoding are examples of this agent. 
We make the following assumption about the inference algorithm.\footnote{We empirically verify that this assumption holds for the agents we evaluate with $N = 10$ and $\gamma$ ranging from 0.7 to 0.9. We estimate $\gamma$ by computing the fraction of evaluation examples where the agent's model ranks $\agentinstr_{\textrm{infer}}$ above $N$ samples drawn from it. } 

\begin{assumption*}[Better-than-sampling inference algorithm]
We assume the inference algorithm is better at finding $\argmax_{\instr \in \mathcal{U}} \agentspeaker(\instr \mid \agenttraj)$ than drawing a small number of $N$ samples from $\agentspeaker$.
Formally, let $\gamma$ be the probability of drawing $\agenttraj$ and a set of $N$ instructions from $\agentspeaker$ such that $\agentspeaker(\agentinstr_{\textrm{infer}} \mid \agenttraj) > \max_{\instr \in \gU_{\textrm{sub}}} \agentspeaker(\instr \mid \agenttraj)$, where $\agentinstr_{\textrm{infer}} = \textsc{GEN}(\agentspeaker, \agenttraj)$. We assume that $\gamma$ is large for a small integer $N > 0$.
\end{assumption*}

If this assumption holds, then with high probability, the agent $r$ behaves identically to a bounded pragmatic speaker that computes its output as:\looseness=-1
\begin{align}
\agentinstr &\triangleq \argmax_{\instr \in \mathcal{U}_{\textrm{sub}}^r} \agentspeaker(\instr \mid \agenttraj) \\
\mathcal{U}_{\textrm{sub}}^r &\triangleq \{ \agentinstr_{\textrm{infer}} \} \cup \{ \instr_i \sim \agentspeaker \mid 1 \leq i \leq N \}
\label{eqn:agent_candidate_set}
\end{align} This agent uses $\agentspeaker$ as both the base speaker $\basespeaker$ and ToM listener $\tomlistener$.
Due to our assumption, on most inputs, the agent outputs $\agentinstr_\textrm{infer}$, similar to the original agent.
We employ this bounded pragmatic speaker as the proxy for the original agent in comparisons with humans, and also refer to it as $r$. 

\paragraph{Evaluation scheme.}
To evaluate a cognitive capability (search or pragmatic) of a speaker $r$, we compute the performance gap between it and an oracle agent that is at human level on the evaluated capability, but is equally good as it is at the other capability.
Specifically, we define $r_{\textrm{search}}^{\star}$ to be an oracle speaker that employs $\agentspeaker$ as the ToM model but is given a ``gold'' candidate set $\mathcal{U}^{\star}_{\textrm{cand}}$ that always contains a human-generated reference instruction $\instr^{\star}$. 
It selects its output as follows
\begin{align}
    \instr_{\textrm{search}}^{\star} 
    &\triangleq \argmax_{\instr \in \mathcal{U}^{\star}_{\textrm{cand}}} \agentspeaker(\instr \mid \agenttraj)
\label{eqn:oracle_search}
\end{align} This agent has similar pragmatic capability as $r$ but human-level search capability. 
Next, we construct $r_{\textrm{pragmatic}}^{\star}$, an oracle that generates candidates using $\agentspeaker$ but employs a human $\humanlistener$ to rank the candidates\looseness=-1
\begin{align}
    &\instr_{\textrm{pragmatic}}^{\star} 
    \triangleq \argmax_{\instr \in \mathcal{U}_{\textrm{sub}}^r} \humanlistener(\agenttraj \mid \instr) 
\label{eqn:oracle_tom}
\end{align} with $\mathcal{U}_{\textrm{sub}}^r$ from \autoref{eqn:agent_candidate_set}. 
The search capability of $r_{\textrm{pragmatic}}^{\star}$ is as good as $r$ but its pragmatic capability is that of a human. 

We calculate the \textit{prospective performance gain} (PPG) with respect to each capability as follows 
\begin{align}
\ppgsearch(r) &\triangleq \perf(r_{\textrm{search}}^{\star}) - \perf(r) \\
\ppgprag(r) &\triangleq \perf(r_{\textrm{pragmatic}}^{\star}) - \perf(r)
\end{align} where $\rho$ is the performance on held-out data (\autoref{eqn:perf_metric} of \autoref{sec:problem}). 
Each metric computes the potential improvement if the corresponding capability is upgraded to match with that of a human. 
Comparing the two metrics reveals which of the two capabilities of $r$ is currently more deficient and informs future development direction for the agent.
For example, if $\ppgsearch(r)$ is large and $\ppgprag(r)$ is small, it means that $r$ is scoring the candidate instructions highly accurately but it is bad at finding high-score instructions. 
In this case, developers may want to focus on devising a more effective inference algorithm to improve the search capability of $r$. 
On the other hand, if $r$ estimates poorly calibrated scores, signified by $\ppgprag(r)$ being large, enhancing its inference algorithm is fruitless, but endowing it with a module that simulates the listener's behavior more accurately would boost its performance.

\section{Improving Pragmatic Capability with Ensemble Instruction-Following Agents}
\label{sec:improve_tom}

In cases where our evaluation scheme indicates that the pragmatic capability of a language model is deficient, we improve it by installing a better ToM listener model.
A common approach to learning this listener model is to use the same dataset used for learning the speaker model \cite{andreas2016reasoning,fried2017unified,fried2018speaker}. 
We argue that this approach has a potential drawback.
A ToM model learned in this way is only exposed to human-generated input instructions.
At deployment time, it would likely experience a \textit{covariate shift} because as a ToM model, the model is then asked to score instructions generated by a speaker model, not by humans.
These instructions may be incorrect, ungrammatical, or may simply have a different style than human-generated instructions.
This covariate shift would hamper the model's judgement.
Our preliminary experiments (Appendix \autoref{appendix_single_ensemble_listners}) confirm that using a listener trained on only human-generated inputs as the ToM model hurts rather than improves the performance of various speakers. 

We show that this problem can be alleviated by employing ToM models that have calibrated uncertainty on unseen instructions.
We obtain calibrated models through ensembling \cite{lakshminarayanan2017simple}:
we train listener models $\hat L^{(k)}(\traj \mid  \instr )$, $k=1\dots{}K$, each on a random $90\%$ subset of the training data with different random initial seeds.

We also leverage access to a simulation of the environment to construct better ToM models. 
Note that the probability that a ToM model $\tomlistener$ assigns to an instruction can be seen as an expectation of a binary metric: $\tomlistener(\agenttraj \mid \instr) = \mathbb{E}_{\traj \sim \tomlistener(\cdot \mid \instr)}\left[ \mathds{1}\{ \traj = \agenttraj \} \right]$, which does not award credit if $\traj$ overlaps only partially with $\agenttraj$.
We propose two augmentations: (i) replace the binary metric with a soft metric $\Psi(\traj , \agenttraj)$ that can measure partial similarity between trajectories and (ii) approximate the expectation by executing listeners $\hat L^{(k)}$ in the simulated environment to sample trajectories.  
Our final model selects its instruction as:\looseness=-1
\begin{align}
    &\agentinstr_{\textrm{augment-ToM}} 
    \triangleq \argmax_{\instr \in \mathcal{U}_{\textrm{sub}}^r} \tomlistener(\agenttraj \mid \instr) \\
    & \tomlistener(\agenttraj \mid \instr) \propto \frac{1}{KM} \sum_{k = 1}^K \sum_{j = 1}^M \Psi(\traj_{j}(\hat L^{(k)}, \instr), \agenttraj) \nonumber \\
    &\mathcal{U}_{\textrm{sub}}^r \triangleq \{ \agentinstr_{\textrm{infer}} \} \cup \{ \instr_i \sim \agentspeaker \mid 1 \leq i \leq N \} \nonumber
\label{eqn:improve_tom}
\end{align} where $\traj(L, \instr)$ denotes a trajectory sampled from a listener model $L$ conditioned on an instruction $\instr$, and $M$ is the number of trajectories we sample from each listener. 
Essentially, the score $\tomlistener( \agenttraj \mid \instr)$ of each candidate instruction is the average performance metric of $K$ listeners, each of which attempts to follow the instruction $M$ times.

\section{Experimental Setup}
\label{sec:exp_setup}

\paragraph{Environment and Dataset.}
We employ Matterport3D \cite{anderson2018vision}, a photo-realistic simulator of the visual perception of a person walking in an indoor environment. 
At any location, an agent is provided with RGB images capturing the 360-degree panoramic view when looking from that location. 

We train and evaluate our models using the Room-to-Room (R2R) language-based navigation dataset. 
Each data point was collected by asking an English-speaking crowd-worker to write a verbal description of a path in an environment. 
The dataset is split into a training set (61 environments, 4,675 paths), a seen validation set (environments seen during training, 340 paths), and an unseen validation set (11 environments unseen during training, 783 paths).
We train the models using the training set and perform model selection on the unseen validation set.
Performance metrics are computed on the seen validation set. \looseness=-1

\paragraph{Speaker Models.}
We evaluate three speaker architectures: (1) a decoder-only \gpt pre-trained on text \cite{radford2019language}; (2) an LSTM encoder-decoder \cite{shen2021much}; (3) a Transformer encoder-decoder \cite{vaswani2017attention}.
Parameters of the latter two models are randomly initialized. Details are in Appendix \autoref{appendix_speaker_models}. \looseness=-1

\paragraph{Human Evaluation.}
\label{sec:human_eval}
We evaluate each speaker model on 75 paths in the seen validation data split. 
In the end, we have annotated 1,200 instructions generated by 16 different systems (humans, 3 speaker models, and their ablated and augmented versions).
To evaluate a speaker model, we present its generated instructions to a human annotator and ask them to follow the instructions to navigate in Matterport3D environments.
We adapt the PanGEA tool\footnote{\url{https://github.com/google-research/pangea}} to setup a web navigation interface and create a task on Amazon Mechanical Turk (MTurk).
We recruit 213 human evaluators in total. 
More details about the setting are given in Appendix \autoref{appendix_human_eval}.

\paragraph{Performance Metrics.} The quality of a speaker is determined by the similarity between the intended trajectories and the actual trajectories that the human evaluators generate by following the speaker's instructions.
We compute these similarity metrics: 
     \textbf{Success rate (\sr)} averages binary indicators of whether the final location of a human-generated trajectory is within three meters of the final location of the intended trajectory;
    \textbf{\spl} \cite{anderson2018evaluation} weights the success indicator with the ratio between the intended traveling distance and the actual one;
    and \textbf{\ndtw and \sdtw} are metrics based on dynamic time-warping alignment \cite{magalhaes2019effective}, capturing the similarity between two point sequences. \ndtw computes only a sequence similarity score while \sdtw weights the score with the success indicator.

\section{Experiments}

We investigate the following questions:
\begin{enumerate}[label=(\alph*)]
\item \textit{How well do the speakers perform on our problem?} We find that, despite implementing advanced architectures, these speakers perform poorly compared to human speakers.  
\item \textit{What causes their performance deficiency?} Using our evaluation scheme, we identify that the speakers possess decent search capability but inadequate pragmatic capability. 
\item \textit{Can we improve the speakers by equipping them with better ToM listeners?} We employ ensembles of state-of-the-art instruction-following agents as ToM listeners for the speakers, and obtain significant improvements. 
\item \textit{What are the challenges in bridging the performance gap with human speakers?} We show that instruction-following agents trained with only human-generated instructions are not optimal for serving as ToM listener models. 
\end{enumerate}

\begin{figure}[t!]
\centering
\includegraphics[width=\linewidth]{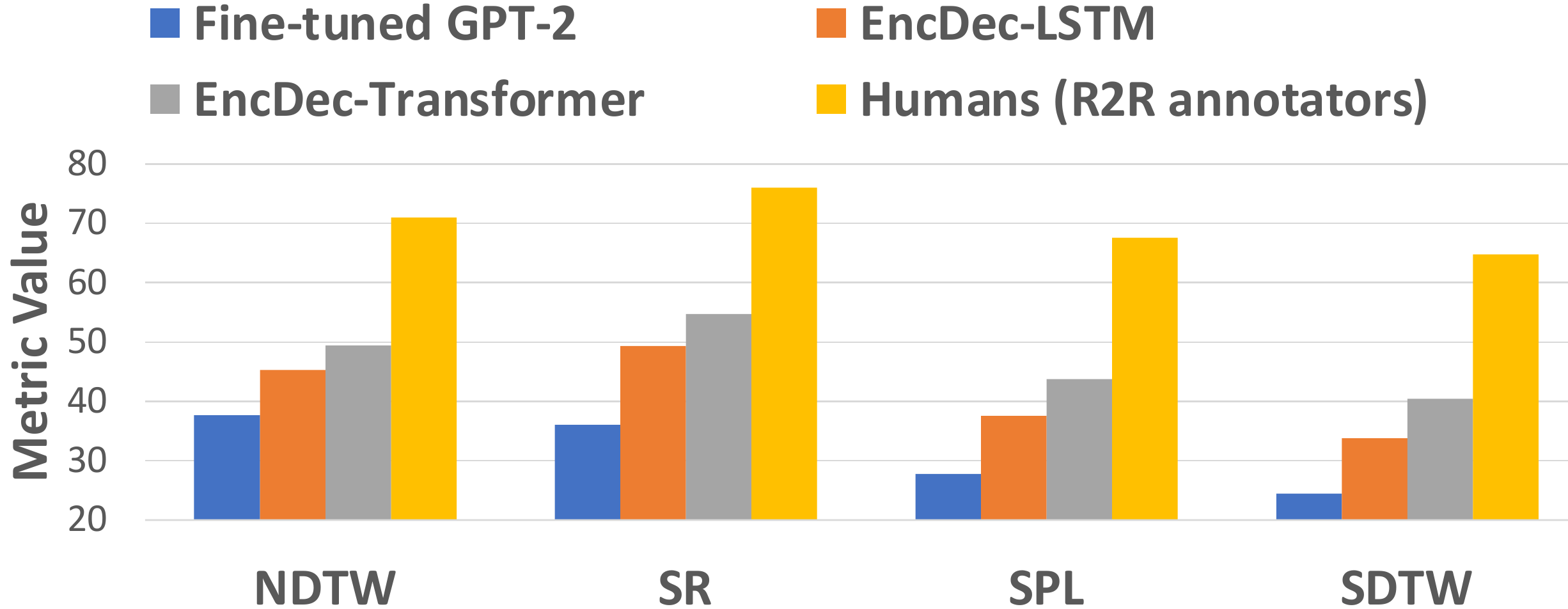}
\caption{Performance of different speakers on held-out evaluation data, grouped by performance metrics (NDTW, SR, SPL, SDTW). Human speakers are annotators of the R2R dataset. There is a considerable gap between model and human speakers.}
\label{fig_speaker_models}
\end{figure}

\paragraph{How well do the speakers perform on our problem?}
As seen in \autoref{fig_speaker_models}, there is a wide margin between the agent speakers and the human instructors. 
The best model speaker (\encdectrans) lags behind the humans by 21.6 NDTW points.
The encoder-decoder architecture with cross-attention of \encdectrans outperforms the decoder-only self-attention architecture of \gpt (+11.7 NDTW), indicating that fusing the vision and language features too early in an architecture may be detrimental.
On the other hand, \encdectrans leads over \encdeclstm by 4.1 points NDTW, suggesting that the Transformer architecture is more effective than LSTM in this problem. 
\looseness=-1

\begin{figure}[t!]
\centering
\includegraphics[width=\linewidth]{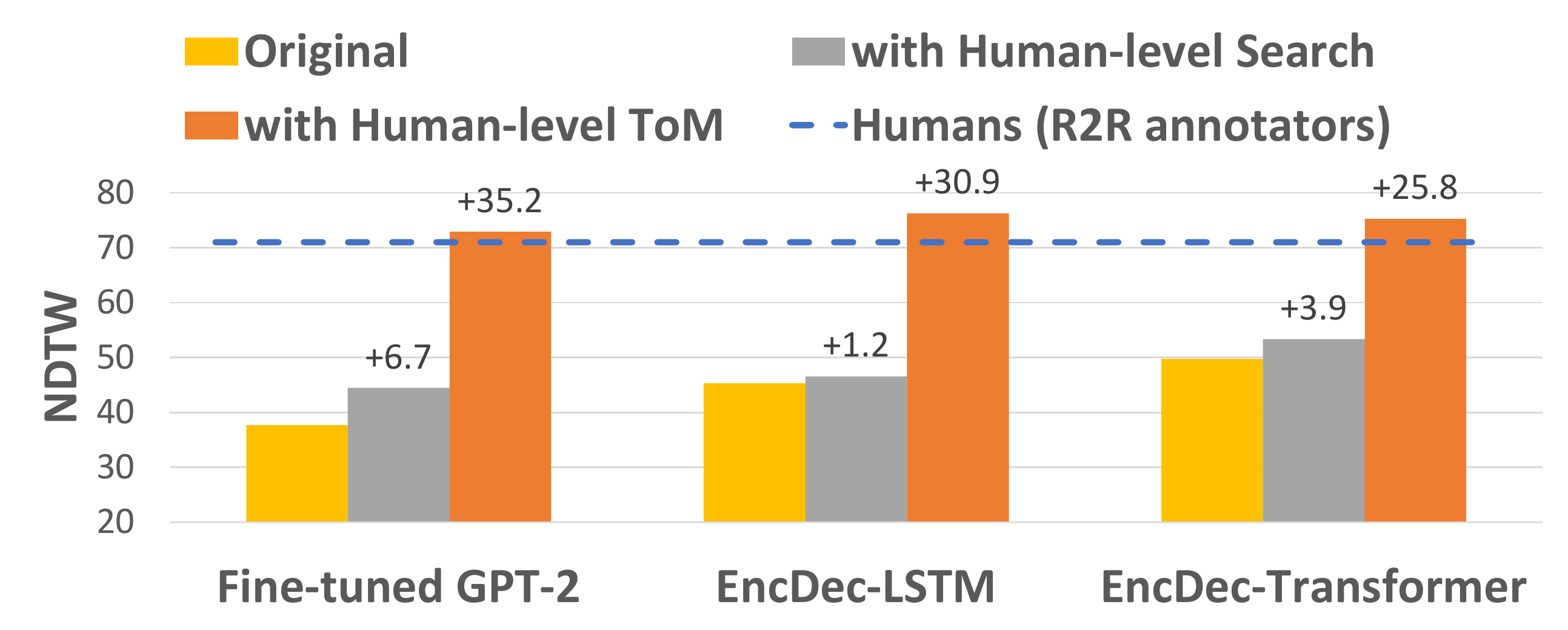}
\caption{Performance (in NDTW) of the speakers and their human-augmented versions. Possessing human-level pragmatic capability improves performance of the speakers, showing that their original pragmatic capability is highly deficient compared to that of a human.\looseness=-1}
\label{fig_oracles}
\end{figure}

\begin{table*}[t!]
\small
\centering
\begin{tabular}{lccc}
    \toprule
    & \multicolumn{3}{c}{Base speaker $\basespeaker$} \\
    ToM listener $\tomlistener$ & Fine-tuned \gpt & \encdeclstm & \encdectrans \\
    \cmidrule(lr){1-1} \cmidrule(lr){2-2} \cmidrule(lr){3-3} \cmidrule(lr){4-4} 
    None & 37.7 ~~\noneperf{0.0} & 45.3 ~~\noneperf{0.0} & 49.4 ~~\noneperf{0.0} \\
    Single VLN-BERT \cite{majumdar2020improving} & 38.9 ~~\upperf{1.2} & 39.8 ~~\downperf{5.5} & 46.2 ~\downperf{3.2} \\
    Ensemble of 10 \envdrop \cite{shen2021much} & 37.8 ~~\upperf{0.1} & ~53.1$^{\dagger}$ \upperf{7.8} & 57.3$^{\dagger}$ \upperf{7.9} \\
    Ensemble of 10 \recurrentbert \cite{hong2021vln} & 43.4 ~~\upperf{5.7} & ~~56.4$^{\ddagger}$ \upperf{11.1} & 54.2 ~~\upperf{4.8} \\
    Humans (skyline) & ~~72.9$^{\ddagger}$ \upperf{35.2} & ~~76.2$^{\ddagger}$ \upperf{30.9} & ~~75.2$^{\ddagger}$ \upperf{25.8} \\
    \bottomrule
\end{tabular}
\caption{Performance (in NDTW) of the speakers when equipped with different ToM models. Each base speaker generates 11 candidates (i.e. $N$ = 10). Ensemble listeners significantly improve performance.
$^{\ddagger}$ and $^{\dagger}$ indicate results that are significantly higher than those of ``None'' (row 1) with $p < 0.05$ and $p < 0.1$, respectively (two-related-sample t-test). \looseness=-1}
\label{tab:improve_tom}
\end{table*}

\begin{table*}[t!]
\small
\centering
\begin{tabular}{lccc}
    \toprule
     & \multicolumn{3}{c}{Listener} \\
    Instructions generated by & \vlnbert & \envdrop & \recurrentbert \\
    \midrule
    Humans (R2R dataset) & 65.4 ~\noneperfdown{0.0} & 47.2 ~\noneperfdown{0.0} & 65.0 ~\noneperfdown{0.0}  \\
    Fine-tuned \gpt & ~~43.1$^{\ddagger}$ \downperf{22.3} & ~~~31.6$^{\ddagger}$ \downperf{15.6} & ~39.9$^{\ddagger}$ \downperf{25.1} \\
    \encdeclstm & ~50.0$^{\ddagger}$ \downperf{15.4} & ~43.7 ~\downperf{3.5} & ~49.3$^{\ddagger}$ ~\downperf{15.7} \\
    \encdectrans & ~52.1$^{\ddagger}$ \downperf{13.3} & ~41.5 ~\downperf{5.5} & ~51.9$^{\ddagger}$ \downperf{13.1} \\
    \bottomrule
\end{tabular}
\caption{Agreement (in NDTW) of human and model listeners on instructions generated by different speakers. The level of agreement decreases substantially when shifting from human-generated to model-generated instructions. 
$^{\ddagger}$ indicate results that are significantly lower than the human skyline (row 1) with $p < 0.05$ (according to a two-related-sample t-test).\looseness=-1}
\label{tab:covariate_shift}
\end{table*}

\paragraph{What causes the speakers' deficiency?} 
Next, we investigate whether the lack of search or pragmatic capability is responsible for the deficiency of the speakers. 
The prospective performance gains presented in \autoref{fig_oracles} show that it is under-performed pragmatic capability that primarily causes the models to perform poorly. 
Specifically, while equipping the models with oracle search capability only improves their performance by 9.4\% on average, granting them oracle pragmatic capability nearly doubles their performance metrics. 
In fact, the search capability of the models is already as good as that of the humans we employ, because the models with oracle pragmatic capability achieve even slightly higher NDTW scores than the human speakers. 

\begin{figure*}[t!]
\centering
\includegraphics[width=\linewidth]{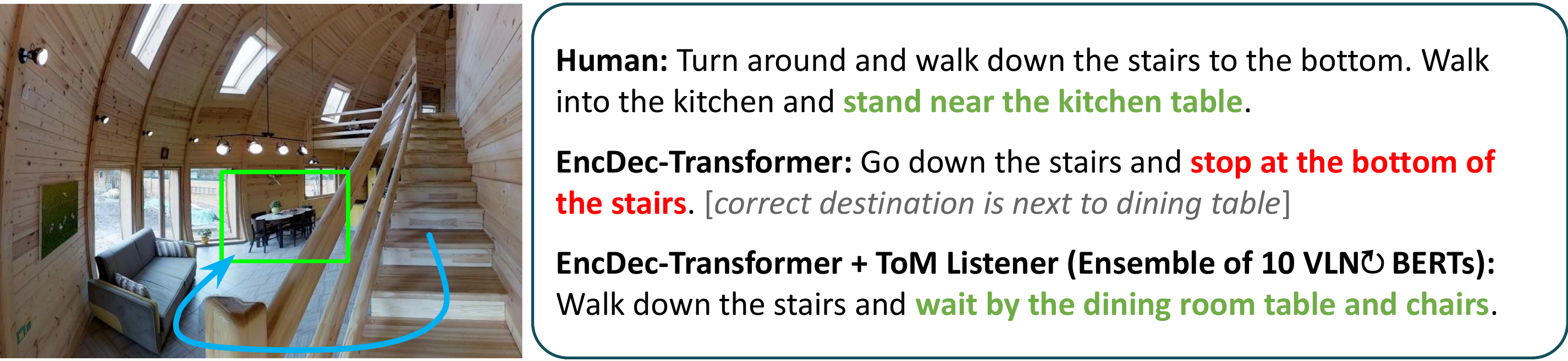}
\caption{A qualitative example where the pragmatic speaker (the last model) avoids missing information by simulating the interpretation of the human listener. \looseness=-1}
\label{fig_qualitative}
\end{figure*}

\paragraph{Can we improve the speakers by equipping them with better ToM models?}
Following the procedure described in Section \autoref{sec:improve_tom}, we train  state-of-the-art instruction-following agents to serve as ToM listener models for the speakers. 
Performances of different combinations of speakers and listeners are given in \autoref{tab:improve_tom}.
We see the largest improvement (+7.9 NDTW) over the best base speaker (\encdectrans) by augmenting this speaker with an ensemble of 10 \envdrop listeners as the ToM model. 
In \autoref{fig_qualitative}, we show a qualitative example where having a ToM listener enables the speaker to generate a more accurate instruction. More examples are shown in Appendix \autoref{appendix_qualitative examples}.

We observe that ensemble models consistently outperform single models.
More results about the effectiveness of ensemble listeners compared to single listeners are given in Appendix \autoref{appendix_single_ensemble_listners}. \looseness=-1

\paragraph{What are the challenges in bridging the performance gap with human speakers?}
Despite the promising improvements, there remains a large gap of 17.9 NDTW points between our best speaker and the human speakers.
As suggested by \autoref{fig_oracles}, this gap can be closed by developing accurate ToM models. 
We argue that optimal ToM models cannot be simply obtained by learning optimal instruction-following agents, because the latter is learned to execute \textit{human-generated} instructions while the former is asked to rank \textit{model-generated} instructions. 
To illustrate the difference, we measure the agreement between human and model listeners on instructions generated by different speakers. 
We define the agreement score between a human $L_h$ and a model $\hat L$ as
\begin{align}
&\textrm{Agreement}(L_h, \hat L) \nonumber \\
&= \textrm{Average}_{\instr \in \mathcal{D}_{\textrm{eval}}} \left( \textrm{NDTW}(\traj_{h}(\instr), \hat \traj(\instr)) \right) 
\end{align} where $\traj_{h}(\instr)$ and $\hat \traj(\instr)$ are the trajectories generated by $L_h$ and $\hat L$ given $\instr$, respectively, and $\mathcal{D}_{\textrm{eval}}$ denotes the R2R seen validation set. 

As seen in \autoref{tab:covariate_shift}, the listener agents agree more with the humans on human-generated instructions than on model-generated ones. 
The results imply even an optimal instruction-following agent can fail to improve a base speaker in the presence of an input distribution mismatch.  
We thus advocate for developing ToM models that are robust or can adapt quickly against covariate shift, and for evaluating performance of these models on model-generated instructions.

\section{Conclusion}
This work introduces a framework for analyzing task-oriented cognitive capabilities of instruction-generation language models. 
We show that insights from the analysis are helpful in directing development on these models. 
Our results highlight the necessity of constructing better ToM models for improving these models. 
We argue that learning accurate ToM listener models is met with novel, distinct challenges. 
We hope that our findings will motivate the community to focus more on evaluating task-oriented cognitive capabilities and to create datasets, training methods, and evaluation procedures for enhancing the pragmatic capability of language models.

\section*{Limitations}

Our work is predicated on hypothetical models of human cognition. 
These models are still under development by cognitive scientists and need to be validated in more realistic domains.
Our method assumes access to a simulation of the environment, which may be costly to construct in some domains. 

In general, instruction generation agents pose substantial risk to humans. 
Previous studies have shown that humans can become overly reliant on AI instructions and commit disastrous mistakes \cite{robinette2016overtrust}.
It is thus important for practitioners to comprehend the constraints of our experimental setting.
Our experiments take place in a coarse simulator of real-world indoor environments, which restricts the action and perception of the human listeners. 
Due to the expensive cost and the large number of agent variants, our human evaluation remains limited in terms of population scale and diversity, and the comprehensiveness of the questionnaires. 
As each instruction is only evaluated by a single human, we have not investigated the variance of the interpretation of the same instruction among different humans.
In addition, human evaluators may ``guess'' a path even if a part of the instruction is misleading or impossible to follow. 
Hence, the path-similarity metrics may not reflect faithfully the quality of the instructions.
Nevertheless, results shown in \autoref{tab:app_main_results} of \autoref{appendix_human_eval} indicates that instructions generated by our agents are almost as easy to interpret as those generated by humans.
But again, these results are still subject to the constraints of our annotator population.
To deploy our method, practitioners should carefully re-evaluate its safety and effectiveness in conditions that closely emulate the deployment conditions. 
\section*{Acknowledgments}
We thank Jieyu Zhao, Trista Cao and our reviewers for providing helpful feedback to improve the manuscript.
We are also grateful to Patrick Shafto and Ted Sumers for great discussions and references to mathematical models of human cognition. 
\looseness=-1

\bibliography{journal_full,anthology,custom}

\begin{thebibliography}{75}
\expandafter\ifx\csname natexlab\endcsname\relax\def\natexlab#1{#1}\fi

\bibitem[{Anderson et~al.(1991)Anderson, Bader, Bard, Boyle, Doherty, Garrod,
  Isard, Kowtko, McAllister, Miller et~al.}]{anderson1991hcrc}
Anne~H Anderson, Miles Bader, Ellen~Gurman Bard, Elizabeth Boyle, Gwyneth
  Doherty, Simon Garrod, Stephen Isard, Jacqueline Kowtko, Jan McAllister, Jim
  Miller, et~al. 1991.
\newblock The hcrc map task corpus.
\newblock \emph{Language and speech}, 34(4):351--366.

\bibitem[{Anderson et~al.(2018{\natexlab{a}})Anderson, Chang, Chaplot,
  Dosovitskiy, Gupta, Koltun, Kosecka, Malik, Mottaghi, Savva
  et~al.}]{anderson2018evaluation}
Peter Anderson, Angel Chang, Devendra~Singh Chaplot, Alexey Dosovitskiy,
  Saurabh Gupta, Vladlen Koltun, Jana Kosecka, Jitendra Malik, Roozbeh
  Mottaghi, Manolis Savva, et~al. 2018{\natexlab{a}}.
\newblock On evaluation of embodied navigation agents.
\newblock \emph{arXiv preprint arXiv:1807.06757}.

\bibitem[{Anderson et~al.(2018{\natexlab{b}})Anderson, Wu, Teney, Bruce,
  Johnson, S{\"u}nderhauf, Reid, Gould, and Van
  Den~Hengel}]{anderson2018vision}
Peter Anderson, Qi~Wu, Damien Teney, Jake Bruce, Mark Johnson, Niko
  S{\"u}nderhauf, Ian Reid, Stephen Gould, and Anton Van Den~Hengel.
  2018{\natexlab{b}}.
\newblock Vision-and-language navigation: Interpreting visually-grounded
  navigation instructions in real environments.
\newblock In \emph{Proceedings of the IEEE conference on computer vision and
  pattern recognition}, pages 3674--3683.

\bibitem[{Andreas and Klein(2016)}]{andreas2016reasoning}
Jacob Andreas and Dan Klein. 2016.
\newblock \href {https://doi.org/10.18653/v1/D16-1125} {Reasoning about
  pragmatics with neural listeners and speakers}.
\newblock In \emph{Proceedings of the 2016 Conference on Empirical Methods in
  Natural Language Processing}, pages 1173--1182, Austin, Texas. Association
  for Computational Linguistics.

\bibitem[{Anil et~al.(2023)Anil, Dai, Firat, Johnson, Lepikhin, Passos,
  Shakeri, Taropa, Bailey, Chen et~al.}]{anil2023palm}
Rohan Anil, Andrew~M Dai, Orhan Firat, Melvin Johnson, Dmitry Lepikhin,
  Alexandre Passos, Siamak Shakeri, Emanuel Taropa, Paige Bailey, Zhifeng Chen,
  et~al. 2023.
\newblock Palm 2 technical report.
\newblock \emph{arXiv preprint arXiv:2305.10403}.

\bibitem[{Bao et~al.(2022)Bao, Ghosh, and Chai}]{bao2022learning}
Yuwei Bao, Sayan Ghosh, and Joyce Chai. 2022.
\newblock \href {https://doi.org/10.18653/v1/2022.acl-long.202} {Learning to
  mediate disparities towards pragmatic communication}.
\newblock In \emph{Proceedings of the 60th Annual Meeting of the Association
  for Computational Linguistics (Volume 1: Long Papers)}, pages 2829--2842,
  Dublin, Ireland. Association for Computational Linguistics.

\bibitem[{Baron-Cohen et~al.(1985)Baron-Cohen, Leslie, and
  Frith}]{baron1985does}
Simon Baron-Cohen, Alan~M Leslie, and Uta Frith. 1985.
\newblock Does the autistic child have a “theory of mind”?
\newblock \emph{Cognition}, 21(1):37--46.

\bibitem[{Bender and Koller(2020)}]{bender2020climbing}
Emily~M Bender and Alexander Koller. 2020.
\newblock Climbing towards nlu: On meaning, form, and understanding in the age
  of data.
\newblock In \emph{Proceedings of the 58th annual meeting of the association
  for computational linguistics}, pages 5185--5198.

\bibitem[{Bloom and Fischler(1980)}]{bloom1980completion}
Paul~A Bloom and Ira Fischler. 1980.
\newblock Completion norms for 329 sentence contexts.
\newblock \emph{Memory \& cognition}, 8(6):631--642.

\bibitem[{Bubeck et~al.(2023)Bubeck, Chandrasekaran, Eldan, Gehrke, Horvitz,
  Kamar, Lee, Lee, Li, Lundberg et~al.}]{bubeck2023sparks}
S{\'e}bastien Bubeck, Varun Chandrasekaran, Ronen Eldan, Johannes Gehrke, Eric
  Horvitz, Ece Kamar, Peter Lee, Yin~Tat Lee, Yuanzhi Li, Scott Lundberg,
  et~al. 2023.
\newblock Sparks of artificial general intelligence: Early experiments with
  gpt-4.
\newblock \emph{arXiv preprint arXiv:2303.12712}.

\bibitem[{Byron et~al.(2009)Byron, Koller, Striegnitz, Cassell, Dale, Moore,
  and Oberlander}]{byron2010report}
Donna Byron, Alexander Koller, Kristina Striegnitz, Justine Cassell, Robert
  Dale, Johanna Moore, and Jon Oberlander. 2009.
\newblock \href {https://aclanthology.org/W09-0628} {Report on the {F}irst
  {NLG} {C}hallenge on {G}enerating {I}nstructions in {V}irtual {E}nvironments
  ({GIVE})}.
\newblock In \emph{Proceedings of the 12th {E}uropean Workshop on Natural
  Language Generation ({ENLG} 2009)}, pages 165--173, Athens, Greece.
  Association for Computational Linguistics.

\bibitem[{Call and Tomasello(2011)}]{call2011does}
Josep Call and Michael Tomasello. 2011.
\newblock Does the chimpanzee have a theory of mind? 30 years later.
\newblock \emph{Human Nature and Self Design}, pages 83--96.

\bibitem[{Enrici et~al.(2019)Enrici, Bara, and Adenzato}]{enrici2019theory}
Ivan Enrici, Bruno~G Bara, and Mauro Adenzato. 2019.
\newblock Theory of mind, pragmatics and the brain: Converging evidence for the
  role of intention processing as a core feature ofhuman communication.
\newblock \emph{Pragmatics \& Cognition}, 26(1):5--38.

\bibitem[{FAIR(2022)}]{meta2022human}
FAIR. 2022.
\newblock Human-level play in the game of diplomacy by combining language
  models with strategic reasoning.
\newblock \emph{Science}.

\bibitem[{Frank and Goodman(2012)}]{frank2012predicting}
Michael~C Frank and Noah~D Goodman. 2012.
\newblock Predicting pragmatic reasoning in language games.
\newblock \emph{Science}, 336(6084):998--998.

\bibitem[{Fried et~al.(2018{\natexlab{a}})Fried, Andreas, and
  Klein}]{fried2017unified}
Daniel Fried, Jacob Andreas, and Dan Klein. 2018{\natexlab{a}}.
\newblock \href {https://doi.org/10.18653/v1/N18-1177} {Unified pragmatic
  models for generating and following instructions}.
\newblock In \emph{Proceedings of the 2018 Conference of the North {A}merican
  Chapter of the Association for Computational Linguistics: Human Language
  Technologies, Volume 1 (Long Papers)}, pages 1951--1963, New Orleans,
  Louisiana. Association for Computational Linguistics.

\bibitem[{Fried et~al.(2018{\natexlab{b}})Fried, Hu, Cirik, Rohrbach, Andreas,
  Morency, Berg-Kirkpatrick, Saenko, Klein, and Darrell}]{fried2018speaker}
Daniel Fried, Ronghang Hu, Volkan Cirik, Anna Rohrbach, Jacob Andreas,
  Louis-Philippe Morency, Taylor Berg-Kirkpatrick, Kate Saenko, Dan Klein, and
  Trevor Darrell. 2018{\natexlab{b}}.
\newblock Speaker-follower models for vision-and-language navigation.
\newblock \emph{Advances in Neural Information Processing Systems}, 31.

\bibitem[{Goeddel and Olson(2012)}]{goeddel2012dart}
Robert Goeddel and Edwin Olson. 2012.
\newblock Dart: A particle-based method for generating easy-to-follow
  directions.
\newblock In \emph{2012 IEEE/RSJ International Conference on Intelligent Robots
  and Systems}, pages 1213--1219. IEEE.

\bibitem[{Gold et~al.(2000)Gold, Murray, Bennett, and
  Sekuler}]{gold2000deriving}
Jason~M Gold, Richard~F Murray, Patrick~J Bennett, and Allison~B Sekuler. 2000.
\newblock Deriving behavioural receptive fields for visually completed
  contours.
\newblock \emph{Current Biology}, 10(11):663--666.

\bibitem[{Goodman and Frank(2016)}]{goodman2016pragmatic}
Noah~D Goodman and Michael~C Frank. 2016.
\newblock Pragmatic language interpretation as probabilistic inference.
\newblock \emph{Trends in cognitive sciences}, 20(11):818--829.

\bibitem[{Gopnik and Astington(1988)}]{gopnik1988children}
Alison Gopnik and Janet~W Astington. 1988.
\newblock Children's understanding of representational change and its relation
  to the understanding of false belief and the appearance-reality distinction.
\newblock \emph{Child development}, pages 26--37.

\bibitem[{Grice(1975)}]{grice1975logic}
Herbert~P Grice. 1975.
\newblock Logic and conversation.
\newblock In \emph{Speech acts}, pages 41--58. Brill.

\bibitem[{Haber et~al.(2019)Haber, Baumg{\"a}rtner, Takmaz, Gelderloos, Bruni,
  and Fern{\'a}ndez}]{haber2019photobook}
Janosch Haber, Tim Baumg{\"a}rtner, Ece Takmaz, Lieke Gelderloos, Elia Bruni,
  and Raquel Fern{\'a}ndez. 2019.
\newblock \href {https://doi.org/10.18653/v1/P19-1184} {The {P}hoto{B}ook
  dataset: Building common ground through visually-grounded dialogue}.
\newblock In \emph{Proceedings of the 57th Annual Meeting of the Association
  for Computational Linguistics}, pages 1895--1910, Florence, Italy.
  Association for Computational Linguistics.

\bibitem[{He et~al.(2016)He, Zhang, Ren, and Sun}]{he2016deep}
Kaiming He, Xiangyu Zhang, Shaoqing Ren, and Jian Sun. 2016.
\newblock Deep residual learning for image recognition.
\newblock In \emph{Proceedings of the IEEE conference on computer vision and
  pattern recognition}, pages 770--778.

\bibitem[{Ho et~al.(2016)Ho, Littman, MacGlashan, Cushman, and
  Austerweil}]{ho2016showing}
Mark~K Ho, Michael Littman, James MacGlashan, Fiery Cushman, and Joseph~L
  Austerweil. 2016.
\newblock Showing versus doing: Teaching by demonstration.
\newblock \emph{Advances in neural information processing systems}, 29.

\bibitem[{Hong et~al.(2021)Hong, Wu, Qi, Rodriguez-Opazo, and
  Gould}]{hong2021vln}
Yicong Hong, Qi~Wu, Yuankai Qi, Cristian Rodriguez-Opazo, and Stephen Gould.
  2021.
\newblock Vln bert: A recurrent vision-and-language bert for navigation.
\newblock In \emph{Proceedings of the IEEE/CVF Conference on Computer Vision
  and Pattern Recognition}, pages 1643--1653.

\bibitem[{Hu et~al.(2023)Hu, Floyd, Jouravlev, Fedorenko, and
  Gibson}]{hu2022fine}
Jennifer Hu, Sammy Floyd, Olessia Jouravlev, Evelina Fedorenko, and Edward
  Gibson. 2023.
\newblock \href {https://arxiv.org/abs/2212.06801} {A fine-grained comparison
  of pragmatic language understanding in humans and language models}.
\newblock In \emph{Proceedings of the 61st Annual Meeting of the Association
  for Computational Linguistics}.

\bibitem[{Jain et~al.(2019)Jain, Magalhaes, Ku, Vaswani, Ie, and
  Baldridge}]{jain2019stay}
Vihan Jain, Gabriel Magalhaes, Alexander Ku, Ashish Vaswani, Eugene Ie, and
  Jason Baldridge. 2019.
\newblock \href {https://doi.org/10.18653/v1/P19-1181} {Stay on the path:
  Instruction fidelity in vision-and-language navigation}.
\newblock In \emph{Proceedings of the 57th Annual Meeting of the Association
  for Computational Linguistics}, pages 1862--1872, Florence, Italy.
  Association for Computational Linguistics.

\bibitem[{Kamath et~al.(2023)Kamath, Anderson, Wang, Koh, Ku, Waters, Yang,
  Baldridge, and Parekh}]{kamath2022new}
Aishwarya Kamath, Peter Anderson, Su~Wang, Jing~Yu Koh, Alexander Ku, Austin
  Waters, Yinfei Yang, Jason Baldridge, and Zarana Parekh. 2023.
\newblock A new path: Scaling vision-and-language navigation with synthetic
  instructions and imitation learning.
\newblock In \emph{Proceedings of the IEEE/CVF conference on computer vision
  and pattern recognition}, pages 10813--10823.

\bibitem[{Koller et~al.(2010)Koller, Striegnitz, Gargett, Byron, Cassell, Dale,
  Moore, and Oberlander}]{koller2010report}
Alexander Koller, Kristina Striegnitz, Andrew Gargett, Donna Byron, Justine
  Cassell, Robert Dale, Johanna~D Moore, and Jon Oberlander. 2010.
\newblock Report on the second nlg challenge on generating instructions in
  virtual environments (give-2).
\newblock In \emph{Proceedings of the 6th international natural language
  generation conference}. The Association for Computer Linguistics.

\bibitem[{Kosinski(2023)}]{kosinski2023theory}
Michal Kosinski. 2023.
\newblock Theory of mind may have spontaneously emerged in large language
  models.
\newblock \emph{arXiv preprint arXiv:2302.02083}.

\bibitem[{Ku et~al.(2020)Ku, Anderson, Patel, Ie, and Baldridge}]{ku2020room}
Alexander Ku, Peter Anderson, Roma Patel, Eugene Ie, and Jason Baldridge. 2020.
\newblock \href {https://doi.org/10.18653/v1/2020.emnlp-main.356}
  {Room-across-room: Multilingual vision-and-language navigation with dense
  spatiotemporal grounding}.
\newblock In \emph{Proceedings of the 2020 Conference on Empirical Methods in
  Natural Language Processing (EMNLP)}, pages 4392--4412, Online. Association
  for Computational Linguistics.

\bibitem[{Lachmy et~al.(2022)Lachmy, Pyatkin, Manevich, and
  Tsarfaty}]{lachmy2021draw}
Royi Lachmy, Valentina Pyatkin, Avshalom Manevich, and Reut Tsarfaty. 2022.
\newblock \href {https://doi.org/10.1162/tacl_a_00522} {{Draw Me a Flower:
  Processing and Grounding Abstraction in Natural Language}}.
\newblock \emph{Transactions of the Association for Computational Linguistics},
  10:1341--1356.

\bibitem[{Lakshminarayanan et~al.(2017)Lakshminarayanan, Pritzel, and
  Blundell}]{lakshminarayanan2017simple}
Balaji Lakshminarayanan, Alexander Pritzel, and Charles Blundell. 2017.
\newblock Simple and scalable predictive uncertainty estimation using deep
  ensembles.
\newblock \emph{Advances in neural information processing systems}, 30.

\bibitem[{Lazaridou et~al.(2020)Lazaridou, Potapenko, and
  Tieleman}]{lazaridou2020multi}
Angeliki Lazaridou, Anna Potapenko, and Olivier Tieleman. 2020.
\newblock \href {https://doi.org/10.18653/v1/2020.acl-main.685} {Multi-agent
  communication meets natural language: Synergies between functional and
  structural language learning}.
\newblock In \emph{Proceedings of the 58th Annual Meeting of the Association
  for Computational Linguistics}, pages 7663--7674, Online. Association for
  Computational Linguistics.

\bibitem[{Le et~al.(2019)Le, Boureau, and Nickel}]{le2019revisiting}
Matthew Le, Y-Lan Boureau, and Maximilian Nickel. 2019.
\newblock Revisiting the evaluation of theory of mind through question
  answering.
\newblock In \emph{Proceedings of the 2019 Conference on Empirical Methods in
  Natural Language Processing and the 9th International Joint Conference on
  Natural Language Processing (EMNLP-IJCNLP)}, pages 5872--5877.

\bibitem[{Levesque et~al.(2012)Levesque, Davis, and
  Morgenstern}]{levesque2012winograd}
Hector Levesque, Ernest Davis, and Leora Morgenstern. 2012.
\newblock The winograd schema challenge.
\newblock In \emph{Thirteenth international conference on the principles of
  knowledge representation and reasoning}.

\bibitem[{Lin et~al.(2022)Lin, Fried, Klein, and Dragan}]{lin2022inferring}
Jessy Lin, Daniel Fried, Dan Klein, and Anca Dragan. 2022.
\newblock \href {https://doi.org/10.18653/v1/2022.acl-long.585} {Inferring
  rewards from language in context}.
\newblock In \emph{Proceedings of the 60th Annual Meeting of the Association
  for Computational Linguistics (Volume 1: Long Papers)}, pages 8546--8560,
  Dublin, Ireland. Association for Computational Linguistics.

\bibitem[{Loshchilov and Hutter(2019)}]{loshchilov2018decoupled}
Ilya Loshchilov and Frank Hutter. 2019.
\newblock \href {https://openreview.net/forum?id=Bkg6RiCqY7} {Decoupled weight
  decay regularization}.
\newblock In \emph{International Conference on Learning Representations}.

\bibitem[{Magalhaes et~al.(2019)Magalhaes, Jain, Ku, Ie, and
  Baldridge}]{magalhaes2019effective}
Gabriel~Ilharco Magalhaes, Vihan Jain, Alexander Ku, Eugene Ie, and Jason
  Baldridge. 2019.
\newblock General evaluation for instruction conditioned navigation using
  dynamic time warping.
\newblock In \emph{NeurIPS Visually Grounded Interaction and Language (ViGIL)
  Workshop}.

\bibitem[{Mahowald et~al.(2023)Mahowald, Ivanova, Blank, Kanwisher, Tenenbaum,
  and Fedorenko}]{mahowald2023dissociating}
Kyle Mahowald, Anna~A Ivanova, Idan~A Blank, Nancy Kanwisher, Joshua~B
  Tenenbaum, and Evelina Fedorenko. 2023.
\newblock Dissociating language and thought in large language models: a
  cognitive perspective.
\newblock \emph{arXiv preprint arXiv:2301.06627}.

\bibitem[{Majumdar et~al.(2020)Majumdar, Shrivastava, Lee, Anderson, Parikh,
  and Batra}]{majumdar2020improving}
Arjun Majumdar, Ayush Shrivastava, Stefan Lee, Peter Anderson, Devi Parikh, and
  Dhruv Batra. 2020.
\newblock Improving vision-and-language navigation with image-text pairs from
  the web.
\newblock In \emph{European Conference on Computer Vision}, pages 259--274.
  Springer.

\bibitem[{Mamassian et~al.(2002)Mamassian, Landy, and
  Maloney}]{mamassian2002bayesian}
Pascal Mamassian, Michael Landy, and Laurence~T Maloney. 2002.
\newblock Bayesian modelling of visual perception.
\newblock \emph{Probabilistic models of the brain}, 13:36.

\bibitem[{Nematzadeh et~al.(2018)Nematzadeh, Burns, Grant, Gopnik, and
  Griffiths}]{nematzadeh-etal-2018-evaluating}
Aida Nematzadeh, Kaylee Burns, Erin Grant, Alison Gopnik, and Tom Griffiths.
  2018.
\newblock \href {https://doi.org/10.18653/v1/D18-1261} {Evaluating theory of
  mind in question answering}.
\newblock In \emph{Proceedings of the 2018 Conference on Empirical Methods in
  Natural Language Processing}, pages 2392--2400, Brussels, Belgium.
  Association for Computational Linguistics.

\bibitem[{Papineni et~al.(2002)Papineni, Roukos, Ward, and
  Zhu}]{papineni-etal-2002-bleu}
Kishore Papineni, Salim Roukos, Todd Ward, and Wei-Jing Zhu. 2002.
\newblock \href {https://doi.org/10.3115/1073083.1073135} {{B}leu: a method for
  automatic evaluation of machine translation}.
\newblock In \emph{Proceedings of the 40th Annual Meeting of the Association
  for Computational Linguistics}, pages 311--318, Philadelphia, Pennsylvania,
  USA. Association for Computational Linguistics.

\bibitem[{Premack and Woodruff(1978)}]{premack1978does}
David Premack and Guy Woodruff. 1978.
\newblock Does the chimpanzee have a theory of mind?
\newblock \emph{Behavioral and brain sciences}, 1(4):515--526.

\bibitem[{Radford et~al.(2021)Radford, Kim, Hallacy, Ramesh, Goh, Agarwal,
  Sastry, Askell, Mishkin, Clark et~al.}]{radford2021learning}
Alec Radford, Jong~Wook Kim, Chris Hallacy, Aditya Ramesh, Gabriel Goh,
  Sandhini Agarwal, Girish Sastry, Amanda Askell, Pamela Mishkin, Jack Clark,
  et~al. 2021.
\newblock Learning transferable visual models from natural language
  supervision.
\newblock In \emph{International Conference on Machine Learning}, pages
  8748--8763. PMLR.

\bibitem[{Radford et~al.(2019)Radford, Wu, Child, Luan, Amodei, Sutskever
  et~al.}]{radford2019language}
Alec Radford, Jeffrey Wu, Rewon Child, David Luan, Dario Amodei, Ilya
  Sutskever, et~al. 2019.
\newblock Language models are unsupervised multitask learners.
\newblock \emph{OpenAI blog}, 1(8):9.

\bibitem[{Robinette et~al.(2016)Robinette, Li, Allen, Howard, and
  Wagner}]{robinette2016overtrust}
Paul Robinette, Wenchen Li, Robert Allen, Ayanna~M Howard, and Alan~R Wagner.
  2016.
\newblock Overtrust of robots in emergency evacuation scenarios.
\newblock In \emph{2016 11th ACM/IEEE international conference on human-robot
  interaction (HRI)}, pages 101--108. IEEE.

\bibitem[{Roman~Roman et~al.(2020)Roman~Roman, Bisk, Thomason, Celikyilmaz, and
  Gao}]{roman2020rmm}
Homero Roman~Roman, Yonatan Bisk, Jesse Thomason, Asli Celikyilmaz, and
  Jianfeng Gao. 2020.
\newblock \href {https://doi.org/10.18653/v1/2020.findings-emnlp.157} {{RMM}: A
  recursive mental model for dialogue navigation}.
\newblock In \emph{Findings of the Association for Computational Linguistics:
  EMNLP 2020}, pages 1732--1745, Online. Association for Computational
  Linguistics.

\bibitem[{Rubio-Fernandez(2021)}]{rubio2021pragmatic}
Paula Rubio-Fernandez. 2021.
\newblock Pragmatic markers: the missing link between language and theory of
  mind.
\newblock \emph{Synthese}, 199(1):1125--1158.

\bibitem[{Sanborn and Chater(2016)}]{sanborn2016bayesian}
Adam~N Sanborn and Nick Chater. 2016.
\newblock Bayesian brains without probabilities.
\newblock \emph{Trends in cognitive sciences}, 20(12):883--893.

\bibitem[{Sanborn et~al.(2010)Sanborn, Griffiths, and
  Navarro}]{sanborn2010rational}
Adam~N Sanborn, Thomas~L Griffiths, and Daniel~J Navarro. 2010.
\newblock Rational approximations to rational models: alternative algorithms
  for category learning.
\newblock \emph{Psychological review}, 117(4):1144.

\bibitem[{Sap et~al.(2022)Sap, Le~Bras, Fried, and Choi}]{sap2022neural}
Maarten Sap, Ronan Le~Bras, Daniel Fried, and Yejin Choi. 2022.
\newblock \href {https://aclanthology.org/2022.emnlp-main.248} {Neural
  theory-of-mind? on the limits of social intelligence in large {LM}s}.
\newblock In \emph{Proceedings of the 2022 Conference on Empirical Methods in
  Natural Language Processing}, pages 3762--3780, Abu Dhabi, United Arab
  Emirates. Association for Computational Linguistics.

\bibitem[{Sap et~al.(2019)Sap, Rashkin, Chen, Le~Bras, and
  Choi}]{sap-etal-2019-social}
Maarten Sap, Hannah Rashkin, Derek Chen, Ronan Le~Bras, and Yejin Choi. 2019.
\newblock \href {https://doi.org/10.18653/v1/D19-1454} {Social {IQ}a:
  Commonsense reasoning about social interactions}.
\newblock In \emph{Proceedings of the 2019 Conference on Empirical Methods in
  Natural Language Processing and the 9th International Joint Conference on
  Natural Language Processing (EMNLP-IJCNLP)}, pages 4463--4473, Hong Kong,
  China. Association for Computational Linguistics.

\bibitem[{Scott-Phillips(2014)}]{scott2014speaking}
Thom Scott-Phillips. 2014.
\newblock \emph{Speaking our minds: Why human communication is different, and
  how language evolved to make it special}.
\newblock Bloomsbury Publishing.

\bibitem[{Shafto et~al.(2014)Shafto, Goodman, and
  Griffiths}]{shafto2014rational}
Patrick Shafto, Noah~D Goodman, and Thomas~L Griffiths. 2014.
\newblock A rational account of pedagogical reasoning: Teaching by, and
  learning from, examples.
\newblock \emph{Cognitive psychology}, 71:55--89.

\bibitem[{Shen et~al.(2022)Shen, Li, Tan, Bansal, Rohrbach, Chang, Yao, and
  Keutzer}]{shen2021much}
Sheng Shen, Liunian~Harold Li, Hao Tan, Mohit Bansal, Anna Rohrbach, Kai-Wei
  Chang, Zhewei Yao, and Kurt Keutzer. 2022.
\newblock How much can clip benefit vision-and-language tasks?
\newblock In \emph{Proceedings of the International Conference on Learning
  Representations}.

\bibitem[{Simon(1957)}]{simon1957models}
Herbert~A Simon. 1957.
\newblock Models of man; social and rational.
\newblock \emph{wiley}.

\bibitem[{Striegnitz et~al.(2011)Striegnitz, Denis, Gargett, Garoufi, Koller,
  and Theune}]{striegnitz2011report}
Kristina Striegnitz, Alexandre~AJ Denis, Andrew Gargett, Konstantina Garoufi,
  Alexander Koller, and Mari{\"e}t Theune. 2011.
\newblock Report on the second second challenge on generating instructions in
  virtual environments (give-2.5).
\newblock In \emph{13th European workshop on natural language generation}.

\bibitem[{Sumers et~al.(2022)Sumers, Hawkins, Ho, Griffiths, and
  Hadfield-Menell}]{sumers2022talk}
Theodore Sumers, Robert~D Hawkins, Mark~K Ho, Thomas~L Griffiths, and Dylan
  Hadfield-Menell. 2022.
\newblock How to talk so ai will learn: Instructions, descriptions, and
  autonomy.
\newblock In \emph{Advances in Neural Information Processing Systems}.

\bibitem[{Talmor et~al.(2019)Talmor, Herzig, Lourie, and
  Berant}]{talmor2018commonsenseqa}
Alon Talmor, Jonathan Herzig, Nicholas Lourie, and Jonathan Berant. 2019.
\newblock \href {https://doi.org/10.18653/v1/N19-1421} {{C}ommonsense{QA}: A
  question answering challenge targeting commonsense knowledge}.
\newblock In \emph{Proceedings of the 2019 Conference of the North {A}merican
  Chapter of the Association for Computational Linguistics: Human Language
  Technologies, Volume 1 (Long and Short Papers)}, pages 4149--4158,
  Minneapolis, Minnesota. Association for Computational Linguistics.

\bibitem[{Tan et~al.(2019)Tan, Yu, and Bansal}]{tan2019learning}
Hao Tan, Licheng Yu, and Mohit Bansal. 2019.
\newblock \href {https://doi.org/10.18653/v1/N19-1268} {Learning to navigate
  unseen environments: Back translation with environmental dropout}.
\newblock In \emph{Proceedings of the 2019 Conference of the North {A}merican
  Chapter of the Association for Computational Linguistics: Human Language
  Technologies, Volume 1 (Long and Short Papers)}, pages 2610--2621,
  Minneapolis, Minnesota. Association for Computational Linguistics.

\bibitem[{Tomasello(2019)}]{tomasello2019becoming}
Michael Tomasello. 2019.
\newblock Becoming human.
\newblock In \emph{Becoming Human}. Harvard University Press.

\bibitem[{Trosborg(2010)}]{trosborg2010pragmatics}
Anna Trosborg. 2010.
\newblock \emph{Pragmatics across languages and cultures}, volume~7.
\newblock De Gruyter Mouton.

\bibitem[{Udagawa and Aizawa(2019)}]{udagawa2019natural}
Takuma Udagawa and Akiko Aizawa. 2019.
\newblock A natural language corpus of common grounding under continuous and
  partially-observable context.
\newblock In \emph{Proceedings of the AAAI Conference on Artificial
  Intelligence}, volume~33, pages 7120--7127.

\bibitem[{Ullman(2023)}]{ullman2023large}
Tomer Ullman. 2023.
\newblock Large language models fail on trivial alterations to theory-of-mind
  tasks.
\newblock \emph{arXiv preprint arXiv:2302.08399}.

\bibitem[{Vaswani et~al.(2017)Vaswani, Shazeer, Parmar, Uszkoreit, Jones,
  Gomez, Kaiser, and Polosukhin}]{vaswani2017attention}
Ashish Vaswani, Noam Shazeer, Niki Parmar, Jakob Uszkoreit, Llion Jones,
  Aidan~N Gomez, {\L}ukasz Kaiser, and Illia Polosukhin. 2017.
\newblock Attention is all you need.
\newblock \emph{Advances in neural information processing systems}, 30.

\bibitem[{Vul et~al.(2014)Vul, Goodman, Griffiths, and Tenenbaum}]{vul2014one}
Edward Vul, Noah Goodman, Thomas~L Griffiths, and Joshua~B Tenenbaum. 2014.
\newblock One and done? optimal decisions from very few samples.
\newblock \emph{Cognitive science}, 38(4):599--637.

\bibitem[{Wang et~al.(2020)Wang, Wang, Paranamana, and
  Shafto}]{wang2020mathematical}
Pei Wang, Junqi Wang, Pushpi Paranamana, and Patrick Shafto. 2020.
\newblock A mathematical theory of cooperative communication.
\newblock \emph{Advances in Neural Information Processing Systems},
  33:17582--17593.

\bibitem[{Wang et~al.(2021)Wang, Montgomery, Orbay, Birodkar, Faust, Gur,
  Jaques, Waters, Baldridge, and Anderson}]{wang2021less}
Su~Wang, Ceslee Montgomery, Jordi Orbay, Vighnesh Birodkar, Aleksandra Faust,
  Izzeddin Gur, Natasha Jaques, Austin Waters, Jason Baldridge, and Peter
  Anderson. 2021.
\newblock Less is more: Generating grounded navigation instructions from
  landmarks.
\newblock \emph{2022 IEEE/CVF Conference on Computer Vision and Pattern
  Recognition (CVPR)}, pages 15407--15417.

\bibitem[{Wimmer and Perner(1983)}]{wimmer1983beliefs}
Heinz Wimmer and Josef Perner. 1983.
\newblock Beliefs about beliefs: Representation and constraining function of
  wrong beliefs in young children's understanding of deception.
\newblock \emph{Cognition}, 13(1):103--128.

\bibitem[{Zellers et~al.(2019)Zellers, Bisk, Farhadi, and
  Choi}]{zellers2019recognition}
Rowan Zellers, Yonatan Bisk, Ali Farhadi, and Yejin Choi. 2019.
\newblock From recognition to cognition: Visual commonsense reasoning.
\newblock In \emph{Proceedings of the IEEE/CVF conference on computer vision
  and pattern recognition}, pages 6720--6731.

\bibitem[{Zhao et~al.(2021)Zhao, Anderson, Jain, Wang, Ku, Baldridge, and
  Ie}]{zhao2021evaluation}
Ming Zhao, Peter Anderson, Vihan Jain, Su~Wang, Alexander Ku, Jason Baldridge,
  and Eugene Ie. 2021.
\newblock On the evaluation of vision-and-language navigation instructions.
\newblock \emph{European Chapter of the Association for Computational
  Linguistics}, pages 1302--1316.

\bibitem[{Zhu et~al.(2021)Zhu, Neubig, and Bisk}]{zhu2021few}
Hao Zhu, Graham Neubig, and Yonatan Bisk. 2021.
\newblock Few-shot language coordination by modeling theory of mind.
\newblock In \emph{International Conference on Machine Learning}, pages
  12901--12911. PMLR.

\end{thebibliography}
\bibliographystyle{acl_natbib}

\clearpage
\appendix

\begin{table}[ht!]
\small
\centering
\begin{tabular}{@{}lllll@{}}
\toprule

Hyperparam          &  \gpt  & Transformer \\ \midrule
Learning rate                &     $10^{-4}$  &     $10^{-4}$  \\
Batch size                   &     4  &     32  \\ 
Optimizer                   &    AdamW   &    AdamW  \\
Num. of training iterations   &  $2 \times 10^{5}$   &  $16 \times 10^{4}$  \\
Max. action steps     &   15  & 35  \\
Max. instruction length      &   100  & 80    \\
Image feature size          &    2048  &    512   \\
Orientation feature size    &    128  &    128  \\
Embedding dropout  &  0.1  & 0.3 \\
Hidden size                 &    768  &    512   \\
Num. of hidden layers     &    1   &  1 \\
Hidden-layer dropout rate   &    0.0  &    0.6   \\
Num. of encoder layers &  -  &  2  \\
Num. of decoder layers &  12  &  2  \\
Transformer dropout rate  &  0.1  & 0.3 \\
Beam size         &      5    &  1    \\ \midrule

\end{tabular}
\caption{Hyperparameters for training the \gpt \encdectrans speakers.}
\label{tab:models_hyperparameters}
\end{table}

\section{Appendices}

\subsection{The Room-to-Room dataset}

The R2R dataset \cite{anderson2018vision} was originally created for training instruction-following agents. 
Each data point was collected by asking a crowd-worker to write a verbal description of a path in an environment. 
In the end, each path was annotated with three instructions.
Each instruction contains 29 words on average.
The dataset is split into a training set (61 environments, 4,675 paths), a seen validation set (340 paths) whose paths are sampled in the training environments, and an unseen validation set (11 environments unseen during training, 783 paths). We do not use the unseen test split because it does not provide ground-truth paths of the descriptions. We use the dataset consistent to their MIT License.

\subsection{Implementation of Speaker Models}
\label{appendix_speaker_models}

We train the speakers with a standard maximum-likelihood objective using the AdamW optimizer \cite{loshchilov2018decoupled} with a learning rate of $10^{-4}$.

The speaker models take a sequence of visual observations and actions from the trajectory $\agenttraj$ as input and output a text instruction $\instr$. The model is trained to estimate conditional probability $S_\theta(\instr | \agenttraj)$. We use grid search to select the model and training hyperparameters, and the best-found values are listed in \autoref{tab:models_hyperparameters}.\looseness=-1

\begin{table*}[ht!]
\centering
\begin{tabular}{lcccccc}
    \toprule
    & \multicolumn{6}{c}{Performance Metrics} \\
    Speaker & SR $\uparrow$ & SPL $\uparrow$ & NDTW $\uparrow$ & SDTW $\uparrow$ & Path Len $\downarrow$ & Interpretability $\uparrow$ \\
    \midrule
    \multicolumn{7}{l}{\textit{Without ToM listener}} \\
    Finetuned \gpt & 36.0 &  27.8  &  37.7  &  24.5  &  20.9 & 2.9\\
    \encdeclstm & 49.3  &  37.6  &  45.3  &  33.8  & 17.4 & 3.3 \\
    \encdectrans & 54.7 &  43.8  &  49.4  & 40.4  &  15.8  & 3.4 \\
    \midrule
    \multicolumn{7}{l}{\textit{With 10 \recurrentbert as ToM listener}} \\
    Finetuned \gpt & 46.7  &  30.9  &  43.4  &  28.1  &  21.2 & 3.0 \\
    \encdeclstm & 54.7  &  46.0  & 56.4   &  41.9  & 14.0  &  3.1 \\
    \encdectrans & 52.0 & 44.0   &  54.2  &  41.6 &  17.7  & 3.2 \\
    \midrule
        Humans (R2R dataset) & 76.0  &  67.6  &  71.0  & 64.8  &  14.2 & 3.6\\
    \bottomrule
\end{tabular}
\caption{Humans evaluation results on instructions generated by the speaker models. The similarity metrics are defined in \autoref{sec:human_eval}. \textit{Path Len} measures the average length of the generated trajectories. \textit{Interpretability} indicates how easy or difficult to follow the instructions according to human evaluators (without knowing the ground-truth trajectory).}
\label{tab:app_main_results}
\end{table*}

\paragraph{Input.} 
The input trajectory $\agenttraj$ is a sequence of panoramic views and actions.
Each panoramic view at time step $t$ is represented by 36 vectors $\{o_{t,i}\}_{i=1}^{36}$, each of which is a visual feature vector extracted from a pre-trained vision model concatenated with orientation features describing the agent's current gaze direction.
The image features of the \gpt model are extracted from a ResNet-152 model \cite{he2016deep}, whereas those of the encoder-decoder models are from a CLIP model \cite{radford2021learning}.
Each ground truth action $a^{\star}_t$, which moves the agent to an adjacent location, is represented by image features from the gaze direction of the agent when looking towards that adjacent location, and orientation features capturing the direction of the adjacent location relative to the agent's current gaze direction. 

\paragraph{Output.}
The output of a speaker model is a language instruction describing the input trajectory. 
At test time, the \gpt model employs beam search, and the encoder-decoder models generate instructions via greedy decoding \cite{shen2021much}.

\paragraph{Training Objective.} We train the speakers with maximum-likelihood objective:
 \begin{equation}
 \max_{\theta} \sum_{(\instr^{\star}, \agenttraj) \in \mathcal{D}_{\textrm{train}}} \sum_{t=1}^{|\instr^{\star}|} \log S_{\theta}(\instr^{\star}_t \mid \agenttraj, \instr^{\star}_{<t})
 \label{equation:training}
 \end{equation}
 where $\theta$ is the speaker model parameters, $\instr^{\star}_t$ is $t$-th word of the ground-truth instruction, and $\instr^{\star}_{<t}$ is the first $t - 1$ words of the instruction.

We select the best model based on the unseen-validation BLEU score \cite{papineni-etal-2002-bleu} of the model-generated instructions with the respect to the ground-truth instructions. 

\paragraph{Tools.} We use SacreBLEU 2.2.1 to compute BLEU scores. For preprocessing and implementing the speaker models, we use Pytorch 1.7.1, NLTK 3.6.7, SentencePiece 0.1.97, and Huggingface Transformers 4.5.1. 

\paragraph{Computation.}
The \gpt model has 124.4 million parameters, and was trained for 24 hours on single NVIDIA GEFORCE RTX 2080 Ti.
The \encdeclstm model has 7.5 million parameters, taking 24 hours to train on single NVIDIA RTX A6000.
The \encdectrans model has 56.6 million parameters, trained on single NVIDIA RTX A6000 for 48 hours.

\subsection{Fine-tuning GPT-2 Speaker Model}

To represent the trajectory features as a sequence of feature vectors to feed into the \gpt model, we first average the view features $\bar o_t$ for each time step:\looseness=-1

\begin{equation}
\bar o_{t} = \frac{1}{36} \sum_{i=1}^{36} o_{t,i}
\end{equation}

We compute the input features $\vec{e}_t^{\star}$ by concatenating the panoramic view features and ground truth action features:

\begin{equation}
\vec{e}_t^{\star} = [\bar o_t; a_t^{\star}]
\end{equation}

The sequence of feature vectors $\vec{e}^{\star}$ representing a trajectory is calculated as follows 
\begin{equation}
\vec{e}^{\star} = [\textrm{tanh}(\vec{e}_1^\star W); \cdots; \textrm{tanh}(\vec{e}_T^\star W)]
\end{equation} where $W$ is parameters of a linear layer. 

For the instruction $\instr^{\star}$, we perform an embedding look-up of its words.
Then, we first prompt the model with $\traj^{\star}$ and then train it to generate $\instr^{\star}$ as a suffix.

\subsection{Training Encoder-Decoder Speaker Models}

Our \encdeclstm model follows the implementation of the speaker in \citet{shen2021much}. 
We implement the \encdectrans model by replacing the LSTM layers of the speaker model described in \citet{tan2019learning} with Transformer layers \cite{vaswani2017attention}.

\subsection{Human Evaluation Interface and Data Collection}
\label{appendix_human_eval}

\begin{figure}[t!]
\centering
\includegraphics[width=0.5\textwidth]{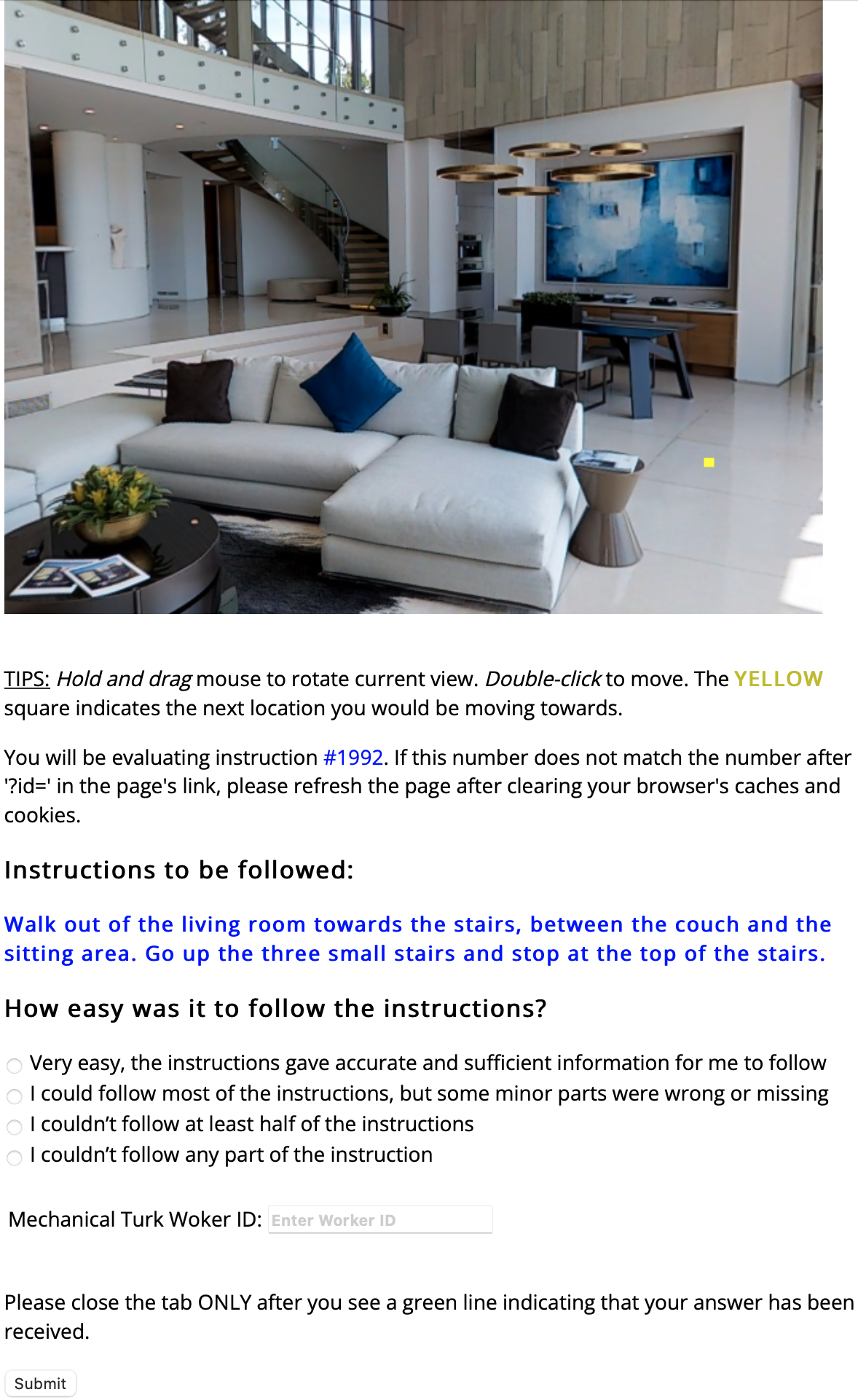}
\caption{Human evaluation interface.}
\label{fig:human_eval}
\end{figure}

\begin{figure}[t!]
\centering
\includegraphics[width=0.5\textwidth]{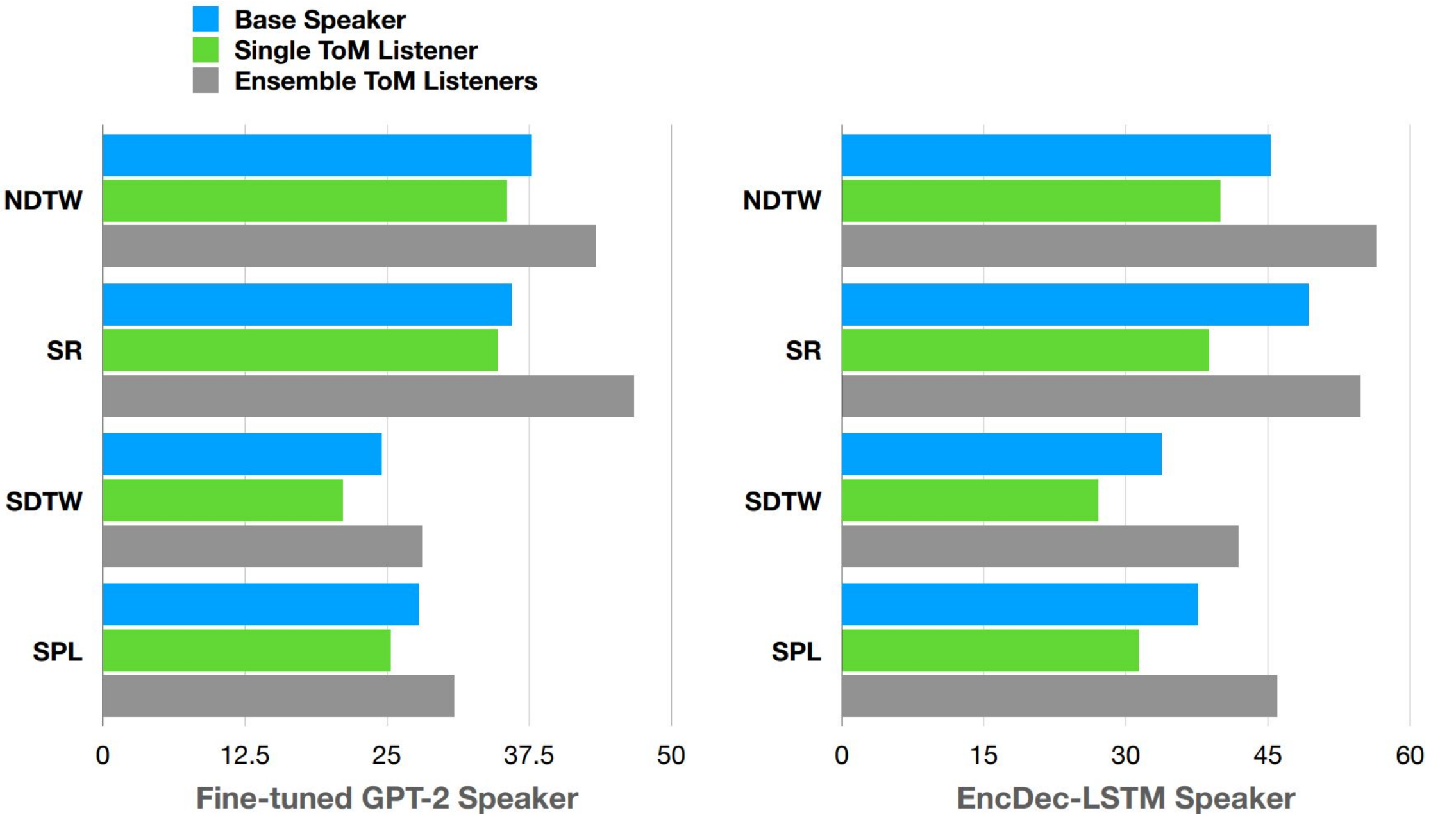}
\caption{Comparison of using single and ensemble ToM listeners.}
\label{fig:single_ensemble_listeners}
\end{figure}

We pay the evaluator \$5.20 per task which takes about 25 minutes, and the payment is decided by state minimum wage. 
For each task, we ask the evaluator to follow six instruction-following sessions. 
One of the six sessions, which appears in all tasks, is a quality-control test featuring an easy-to-follow human-written instruction. 
We only approve an evaluator if they navigate successfully to the goal destination in this test. 
Following \citet{zhao2021evaluation}, we instruct the judges to not explore the environments unnecessarily and not wander back and forth unless they are lost.
We record the trajectories created by the human and use them to compute the performance metrics. 

 \autoref{fig:human_eval} shows the interface for our human evaluation to collect annotations, which we adapted from the PanGEA tool\footnote{\url{https://github.com/google-research/pangea}} consistent with their Apache License v2.0.
After a human evaluator finishes following an instruction, we recorded the path they generate and compute similarity metrics with respect to the ground-truth path. 
After the instruction-following sessions, we ask each evaluator to assess the interpretability of the instructions by asking them how easy (or difficult) it was for them to follow the instruction. 
We provide four rating levels ranging from ``\textit{1: I couldn’t follow any part of the instruction}'' to ``\textit{4: very easy, the instructions gave accurate and sufficient information for me to follow}''. 
The answer of the evaluators is converted to a score between one and four.

\autoref{tab:app_main_results} shows the human evaluation results of the three speaker models we evaluated.

For the human evaluation survey, participants will be restricted to those fluent in English. There are no other restrictions for this study. Participants must be at least 18 years old. Before completing the survey, participants will be shown information about the task requirement: \textit{You are in a building, and are provided with a short set of instructions to navigate to a target location. Please follow the instructions as closely as possible. Do NOT explore the building unnecessarily and do NOT wander back and forth unless you are lost. Please read ALL of the instructions before you start moving.} 

We waive consent for this study for several reasons: 1) Minimal risk: The study collects minimal identifying information and there are no known risks for the subjects beyond everyday computer use. 2) Rights and welfare: All participants will be shown all information regarding task requirements before they complete our survey. They must consent to performing the task before they are shown the questions. 3) Practicality: Since the sessions are conducted online on a large scale, it would be infeasible to require all users to send a signed form. 4) Post participation information: We do not think there is any pertinent information that is not already shared with the participants before or during our experiments, so we do not feel it is necessary to provide any additional information after participation. PI information will be shared with the participants to enable them to obtain additional information about the study post completion.

For data anonymization, we removed the only identifying information, Amazon Mechanical Turk ID, after collecting the human annotation data. This information would also be removed for future dataset release.
The dataset will be released under MIT license terms, which are compatible with those of the tools used to create it, and will be intended for research usage. 

\subsection{Single vs. Ensemble Listeners}
\label{appendix_single_ensemble_listners}

As a preliminary experiment, we compare the effectiveness of a single and an ensemble of 10 \recurrentbert agents when serving as the ToM model of a speaker. 
Results in \autoref{fig:single_ensemble_listeners} show that the ensemble listener is significantly better than the single listener for two different speakers.

\subsection{Qualitative Examples}
\label{appendix_qualitative examples}

\begin{figure*}[t!]
\centering
\begin{subfigure}{\textwidth}
\centering
\includegraphics[width=0.7\linewidth]{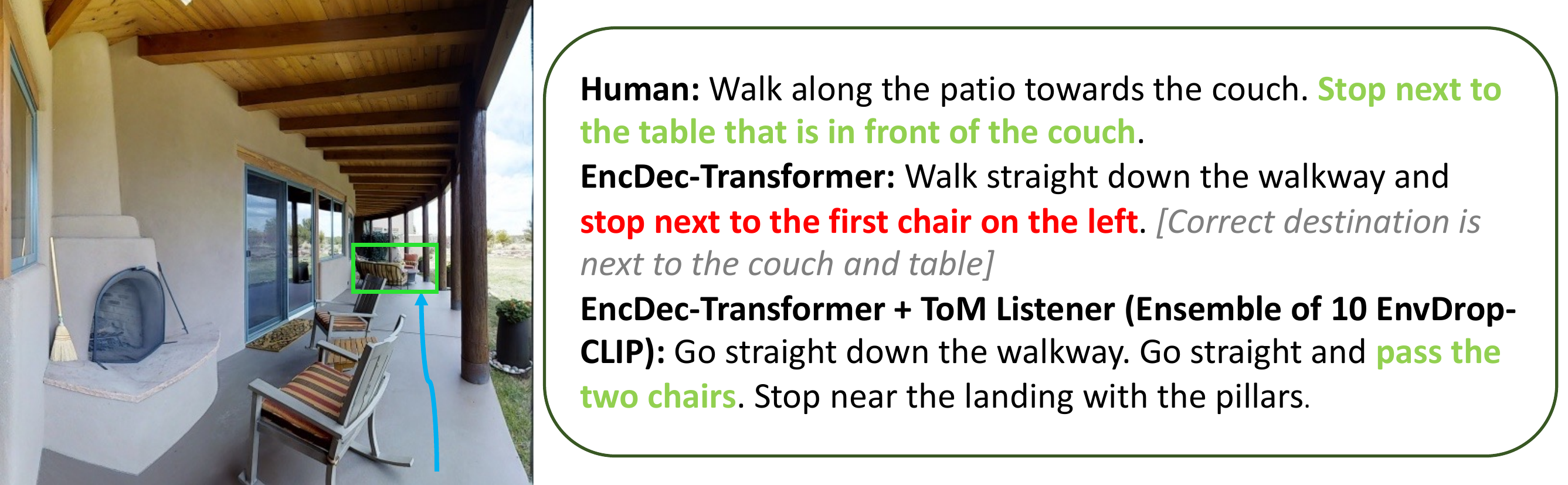}
\caption{}
\end{subfigure}
\hfill
\begin{subfigure}{\textwidth}
\centering
\includegraphics[width=0.7\linewidth]{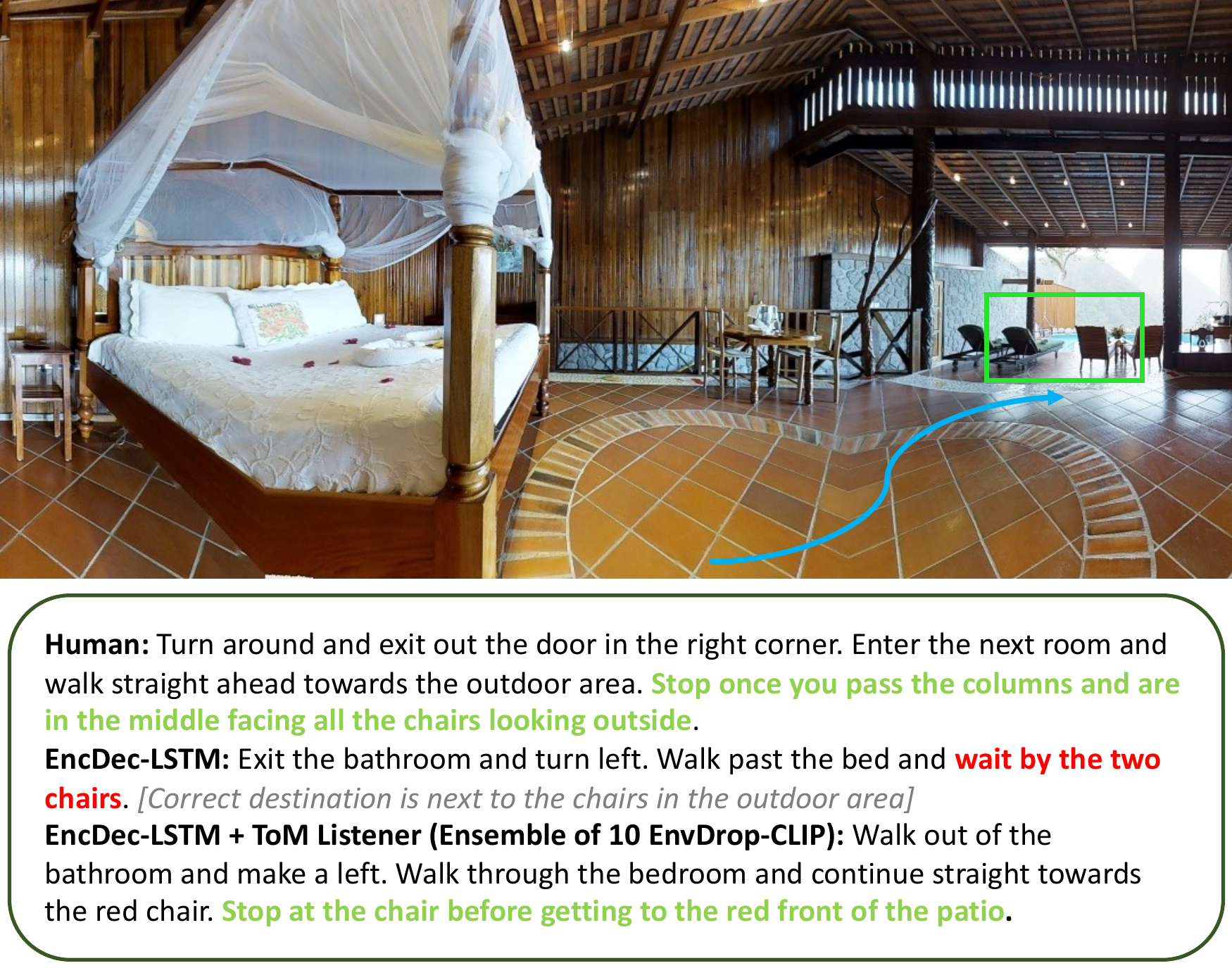}
\caption{}
\end{subfigure}
\caption{Additional qualitative examples where the pragmatic speaker (the last model) avoids missing information by simulating the interpretation of the human listener. \looseness=-1}
\label{fig_qualitative3}
\end{figure*}

In \autoref{fig_qualitative3}, we show additional qualitative examples where having a ToM listener enables the speaker to generate a more accurate instruction.

\end{document}